\documentclass[journal]{IEEEtran}

\usepackage[utf8]{inputenc}
\usepackage{times}
\usepackage{epsfig}
\usepackage{graphicx}
\usepackage{mathrsfs,amsmath,bbm,bm}
\usepackage{breqn}
\usepackage{amssymb}
\usepackage{xcolor}
\usepackage{enumerate}
\usepackage[font=footnotesize,labelfont=bf, tableposition=top]{caption}
\usepackage[font=footnotesize]{subcaption}
\usepackage{float}
\usepackage{footnote}
\usepackage{pifont}
\usepackage{xifthen}
\usepackage{csquotes}
\usepackage{mathtools}

\usepackage{tikz}
\usepackage{pgfplots}
\usepackage{adjustbox}
\usetikzlibrary{calc,positioning}

\usepackage{enumerate}
\usepackage{enumitem}
\usepackage{array}
\usepackage{booktabs}
\usepackage{multirow}
\usepackage{tabularx}
\usepackage{diagbox}
\usepackage{siunitx}
\usepackage{etoolbox}
\robustify\bfseries 
\usepackage{dcolumn}
\newcolumntype{d}[1]{D{.}{.}{#1}}
\makeatletter
\newcommand{\thickhline}{    \noalign {\ifnum 0=`}\fi \hrule height 1pt
    \futurelet \reserved@a \@xhline
}
\newcolumntype{\DC}{@{\hskip\tabcolsep\vrule width 1pt\hskip\tabcolsep}}
\makeatother

\usepackage{hyperref}
\usepackage[numbers,sort&compress]{natbib}

\definecolor{darkgray}{rgb}{.5,0.5,0.5}

\newcommand{\Idea}[1]{{\ifdefined\DRAFT{\color{blue}#1}\fi}}

\newcommand{\Vector}[1]{\mathbf{#1}}
\newcommand{\Matrix}[1]{\mathbf{#1}}
\newcommand{\Tensor}[1]{\mathcal{#1}}

\newcommand{\R}[1]{\mathbb{R}^{#1}}

\newcommand{\argmin}{\operatorname*{argmin}}

\newcommand{\ones}{\bm{\mathbbm{1}}}

\newcommand{\St}{\textrm{\hspace{2mm} subject to \hspace{2mm}}}

\newcommand{\FDF}[1]{\ifthenelse{\isempty{#1}}{\Tensor{D}}{\Matrix{D}_{#1}}}
\newcommand{\AOF}[1]{\ifthenelse{\isempty{#1}}{\boldsymbol{\Phi}}{\boldsymbol{\Phi}_{#1}}}

\newcommand{\SCI}{\Matrix{U}}

\newcommand{\FRT}{\Tensor{F}}
\newcommand{\caFRT}{\Tensor{V}}

\newcommand{\envelope}{(\raisebox{-.5pt}{\scalebox{1.45}{\Letter}}\kern-1.7pt)}

\begin{document}

\title{
Model-based learning of local image features for unsupervised  texture segmentation
}

\author{Martin Kiechle,
        Martin Storath,
        Andreas Weinmann,
        Martin Kleinsteuber
\thanks{M. Kiechle and M. Kleinsteuber are with the Department of Electrical and Computer Engineering, Technical University of Munich, Germany. \{martin.kiechle, kleinsteuber\}@tum.de, www.gol.ei.tum.de}\thanks{M. Storath is with the Image Analysis and Learning Group, Universität Heidelberg, Germany.}\thanks{A. Weinmann is with the Department of Mathematics and Natural Sciences, Darmstadt University of Applied Sciences, and with the Institute of Computational Biology, 
Helmholtz Center Munich, Germany.}}

\maketitle

\begin{abstract}
Features that capture well the textural patterns 
of a certain class of images
are crucial for the performance of texture segmentation methods. The 
manual selection of features  or designing new ones  can be a tedious task.
Therefore, it is desirable to automatically adapt the features to a certain image or class of images.
Typically, this requires a large set of training images with similar textures and ground truth segmentation.
In this work, we propose a framework to learn
features for texture segmentation when no such training data is available.
The cost function for our learning process 
is constructed to match a commonly used segmentation model, the piecewise constant Mumford-Shah model.
This means that the features are learned such that they provide an
approximately piecewise constant feature image 
with a small jump set.
Based on this idea, we develop a two-stage algorithm 
which first learns suitable convolutional features and then performs a segmentation.
We note that the features can be learned  
from a small set of images, from a single image, or even from image patches.
The proposed method achieves a competitive rank in the Prague 
texture segmentation benchmark,
 and it is effective for segmenting histological images.

 \end{abstract}


\IEEEpeerreviewmaketitle

\section{Introduction}

Texture segmentation
is a frequently occurring and challenging problem 
in image processing and computer vision.
For textured images -- such as many natural images \cite{martin2004learning, mobahi2011segmentation},  histological images \cite{mccann2014}, or crystal structures \cite{mevenkamp2016} -- 
the segmentation is typically performed in two stages.
In the first stage, a (vector-valued) feature image is 
derived from the image.
The corresponding features  
are designed to capture 
the local statistical properties or oscillatory patterns of a texture.
Many classical features are based on linear filters \cite{randen1999filtering},
for example Gabor filters \cite{jain1991unsupervised}, 
wavelet frames \cite{unser1995texture},
windowed Fourier transform \cite{azencott1997texture},
followed by a pointwise non-linearity \cite{unser1990nonlinear}. 
Other popular features are based on local spectral histograms \cite{liu2006image}, morphological filters \cite{xia2006morphology},
local statistical descriptors \cite{todorovic2009}
 or local binary patterns \cite{ojala2002multiresolution}.
In the second stage, 
the feature image is segmented.
Popular choices include k-means clustering \cite{jain1991unsupervised, unser1995texture}
or mean shift algorithms \cite{ozden2005image}.
More sophisticated (variational) segmentation models 
additionally enforce spatial regularity of the segment boundaries: 
here, a prominent example 
is the piecewise constant Mumford-Shah model (or Potts model) \cite{geman1984stochastic,mumford1989optimal}; 
it has been used for texture segmentation, for instance in \cite{rousson2003active, kato2006markov, storath2014unsupervised, mevenkamp2016}.

\subsection{Motivation and related work}
Besides the aforementioned works, 
there is a series of more recent contributions
 to unsupervised texture segmentation:
Todorovic and Ahuja create a tessellation of texture super-pixels (texels) and cluster them by a multiscale segmentation and a meanshift algorithm \cite{todorovic2009}. 
Galun et al.~\cite{galun2003} utilize a multiscale aggregation of filter responses and shape elements. 
Haindl and Mikes employ a Gaussian MRF texture model  \cite{haindl2004}
or a 3D auto regressive model \cite{haindl2006},
and they perform segmentation based on a Gaussian mixture model.
Scarpa et al.~\cite{scarpa2009} 
use features based on Markov chains, and then segment by recursively merging them according to their mutual interaction.
Yuan et al.~\cite{yuan2012image} use
local spectral histograms as feature vectors
and formulate the segmentation problem as a multivariate linear regression.
In a follow-up work \cite{yuan2015factorization},  non-negative matrix factorization is used for segmentation.
Storath et al. \cite{storath2014unsupervised}
utilize monogenic curvelets as features
and perform segmentation based on the piecewise constant Mumford-Shah model.
The method of Panagiotakis et al.~\cite{panagiotakis2011,panagiotakis2011slides} is  based on voting of blocks, Bayesian flooding and region merging. Mevenkamp and Berkels \cite{mevenkamp2016} use local Fourier features, which are tailored to images with crystal structures, and segment using a convex relaxation of the piecewise constant Mumford-Shah model.
McCann et al.~\cite{mccann2014}
utilize features derived from local histograms, and segment using nonnegative matrix factorization and image deconvolution.

 It is a fundamental issue that 
  the performance of the features 
  depends strongly on the class of images or even on the single image.
For instance, good features for a natural image may
   perform poorly on a histological image.
  Even more, good features for one natural image may 
  not perform as well on another natural image.
  Thus, the design of the features is a critical task
and there are several approaches to this.
A straightforward idea is to simply 
increase the number of features hoping that
at least some features are well suited for the texture patterns of the processed image.
Unfortunately, the computational effort for segmenting large feature spaces is very high in practice, in particular for segmentation methods which enforce regularity of the boundaries.
To circumvent these problems, a commonly used strategy is
to manually select a subset from the aforementioned larger set of features; 
see for example \cite{kato2006markov, yuan2015factorization}.
\Idea{MEHR CITES}
However, the manual selection requires human supervision 
which typically results in an expensive, time-consuming task.
In principle, for each new class of images one should reevaluate this selection.
To avoid manual design of features for each image or each class of images,
it seems natural to learn them from data.
In a supervised learning setup,
where a sufficiently large training set of images with similar characteristics and a ground truth segmentation is available,
one can use generic methods;
for example the super-pixelation based method of \cite{ren2003learning}
or more recent methods based on convolutional neural networks
\cite{long2015fully}.
In the present unsupervised setup, 
such a training set is not available. 
As a consequence, the challenge is to find a
suitable objective function for the learning task,
and a practical numerical procedure to optimize the features accordingly.

\subsection{Contribution}

In this work, we develop a method
for unsupervised texture segmentation 
where the features are learned from non-annotated data,
i.e., from images without ground truth segmentation.
The main contributions of this work are
\emph{(i)} a model for feature learning of image features 
for texture segmentation in the absence of annotated training data,
and \emph{(ii)} 
a practical algorithm for unsupervised texture segmentation based on that model.

Regarding the basic model \emph{(i)}, 
our starting point is the observation 
that features are often designed such that 
the feature image is approximately constant on a texture segment.
This allows utilizing segmentation algorithms based on a local homogeneity assumption.
The basic idea of our model is to learn 
convolutional features
in a way that they produce approximately piecewise constant feature images. 
Besides reasonable constraints on the filters, such as their norm and mutual coherence,
the objective is to minimize 
the cost function of the popular piecewise constant Mumford-Shah segmentation model, i.e.~the total length of the discontinuity set of the corresponding feature image.
Regarding \emph{(ii)}, 
learning filters based on the proposed model turns out to be a
challenging optimization problem
because it involves a non-smooth and non-convex cost function 
on the (non-convex) unit sphere.
To make it computationally tractable,
we decompose and relax this model:
we obtain the two stages of  filter learning and of segmentation.
For the relaxed learning stage, we employ
a smooth (yet non-convex) approximation of the cost function.
To minimize this cost function, we adapted  
a geometric conjugate gradient descent method proposed in \cite{hawe2013,kiechle2015}
such that it fits with the proposed model.
For the segmentation stage,
we employ the Lagrange formulation of the piecewise constant Mumford-Shah model. 
In particular, implied by the model, we consider a data term based on the Mahalanobis distance.
To solve the corresponding problem,
we extended the approach proposed in \cite{storath2014jump,storath2014fast} in order to be able to deal with the Mahalanobis distance.
Finally, a post-processing as in \cite{yuan2015factorization} merges small spurious regions to large ones.

We evaluate our method on different types of textured images.
A standard benchmark for texture-based segmentation
is the Prague texture segmentation benchmark \cite{haindl2008online}.
Here, our method achieves a top rank.
In particular, the proposed method gives significantly better results than many earlier methods
\cite{todorovic2009, galun2003, haindl2004, haindl2006,scarpa2009, yuan2012image},
and slightly better results than the more recent methods proposed in
\cite{yuan2015factorization,mevenkamp2016}.
 Further, we are competitive with the currently leading method PMCFA \cite{panagiotakis2011,panagiotakis2011slides}.
Besides, our approach 
provides satisfactory segmentation results
on the data set of histological images of \cite{mccann2014}. 
We emphasize that, although this is a quite different image class, only minor adjustments have been necessary.
This shows in particular the flexibility of our method, 
and the potential for segmenting quite different classes of textured images.

\section{A model for unsupervised filter learning for texture segmentation}
As mentioned in the introduction,
our goal is 
to learn suitable features for texture segmentation when no 
 training data with ground truth is available.
Here, we focus on learning convolutional features.
Convolutional filters are a natural choice because they describe the class of linear translation-invariant filters.
A feature image is created by
applying linear filtering  
followed by a (pointwise) nonlinear transform. 
More precisely,
given an image $\SCI \in \mathbb{R}^{M\times N},$
we consider $K$ different convolution filters $\AOF{1}, \ldots, \AOF{K},$ and the resulting filtered images, given in Matlab-type notation by
\[
    \FRT_{:,:,1} = \AOF{1} \SCI,~\ldots,~
    \FRT_{:,:,K} = \AOF{K} \SCI.
\]
In short-hand notation, we write  $\FRT = \AOF{} \SCI.$
Then, to each filter response, the same nonlinear transformation $\sigma$
is applied pixel wise. In general, $\sigma$ is chosen to be symmetric, i.e. $\sigma(x)=\sigma(-x).$
Further, it is required that it has fast decaying slope for large $x$ 
in order to be robust towards outliers in the filter responses. 
The nonlinear transform has proven to be beneficial for texture segmentation:
according to \cite{unser1990nonlinear},
its purpose is to translate differences in dispersion characteristics into differences in mean value.
For further details on choosing $\sigma$ we refer to \cite{unser1990nonlinear}.
In this paper we use a logarithmic non-linearity of the form $\sigma(x):=\log(1+\mu x^2)$ with the free parameter $\mu >0.$
The nonlinear transform is considered to be fixed, 
and we are interested in finding suitable linear convolution operators 
 $\AOF{1}, \ldots, \AOF{K},$ which define the features
 \[
 \Tensor{V} = \sigma(\AOF{} \SCI).
 \]
 Here, $\Tensor{V}$ and $\AOF{} \SCI$ are three dimensional arrays in $\mathbb{R}^{M\times N \times K},$
 and $\sigma$ has to be understood as componentwise application of its scalar version.

Since we are in an unsupervised setup, we have no training data (i.e. no ground truth segmentation) 
for learning the $\AOF{1}, \ldots, \AOF{K}$ from.
In particular, there is no straightforward way to devise 
a loss function for the learning process.
We propose to utilize a loss function 
based on the segmentation model,
which in our case is the piecewise constant Mumford Shah or Potts model:
ideally, the features $\Tensor{V}$ are approximately
constant on the texture, and the segment boundaries are sufficiently regular.
The idea is to learn suitable filters $\AOF{}$ in 
a way such that their responses (after applying the non-linearity) on the segments 
are approximately constant.
We propose to minimize as a cost function 
the length of the discontinuity set of $\Tensor{V},$ 
denoted by $\| \nabla \Tensor{V} \|_0.$ More precisely, we propose as a model 
for choosing the convolution kernels $\AOF{1}, \ldots, \AOF{K},$
\begin{equation}
\label{eq:proposedModel}
\min_{\Tensor{V},\Matrix{\Phi}} \; \| \nabla \Tensor{V} \|_0 \St 
d(\Tensor{V},\sigma(\AOF{} \SCI)) \leq \varepsilon,
\end{equation}
with $\varepsilon>0.$  
Here, the minimum is taken with respect to both $\AOF{},\Tensor{V},$
where the $\AOF{k}$ have unit length, zero mean, and fulfill an incoherence and a certain center condition. (We elaborate on these constraints in Section~\ref{subsec:constraints}.)
The symbol $d$ denotes a metric, in our case the Mahalanobis distance
as explained in Section~\ref{sec:SegmetationStage}.
We note that an optimal pair $\AOF{}^\ast,\Tensor{V}^\ast $ of Eq.~\eqref{eq:proposedModel}
already consists of an optimal filter bank $\AOF{}^\ast$ together with a corresponding optimal segmentation $\Tensor{V}^\ast.$

The model Eq.~\eqref{eq:proposedModel} is computationally hard to access.
In particular, the simultaneous optimization w.r.t.\ both $\Matrix{\Phi}$ and $\Tensor{V}$ is extremely demanding.
As an approximative strategy, we propose a two stage approach as follows.
As a first step, we optimize the filters $\AOF{}$ using a relaxation of Eq.~\eqref{eq:proposedModel} 
as described in Section~\ref{sec:filter_learning}.
For the second step, we notice, that
for fixed $\AOF{},$  the Lagrange form of  Eq.~\eqref{eq:proposedModel} is the piecewise constant Mumford-Shah model.
Therefore, we perform a piecewise constant Mumford-Shah segmentation w.r.t.\ the Mahalanobis distance
(described in Section~\ref{sec:SegmetationStage}) for the obtained feature image.
We note that even this second step of solving the piecewise constant Mumford-Shah problem is known to be an NP hard problem on its own.

For notational brevity, 
we describe the derivation of our method on gray-valued images $\SCI \in \mathbb{R}^{M \times N}$. 
The derivation for multi-channel images follows
the same basic steps. The relevant modifications
regarding the operators $\AOF{}$ and the jump penalty
are described in Section~\ref{sec:vectorvalued}.

\Idea{PRINZIP MIT BILD ILLUSTRIEREN}

\section{Learning stage} \label{sec:filter_learning}

In this section, we discuss how to learn the filters  $\AOF{}$ from a given image.
As a first step, we present a near anisotropic discretization  
of the jump penalty Eq.~\eqref{eq:proposedModel} in Section~\ref{subsec:discretize}.
Then, we relax the model Eq.~\eqref{eq:proposedModel} 
to obtain a computationally better accessible surrogate problem
to perform the learning task in Section~\ref{subsec:relaxLearning}.
Further, we incorporate learning from patch samples in Section~\ref{subsec:learnignFromPatch},
and explain how to deal with the constraints imposed on the filters in Section~\ref{subsec:constraints}, respectively. Then, we sum up the simplified learning problem
and discuss its numerics in Section~\ref{subsec:SimplifiedlearningOpt}.
Finally, we explain how to generalize the approach for multi-channel images
in Section~\ref{subsec:ExtMultiChan}.

As pointed out, we focus on sets of linear filters.
Further, we assume that each filter has a fixed number of $n$ coefficients $\AOF{k} \in \R{n}$.

\subsection{Near isotropic discretization}
\label{subsec:discretize}

First, we deal with a near isotropic discretization of 
the jump penalty $\|\nabla \caFRT  \|_0.$
As in \cite{storath2014fast}, we use 
a finite difference discretization of the form 
\begin{equation}\label{eq:pottsDisc}
	\| \nabla \caFRT \|_0 = \sum_{ s=1}^S  \omega_s \| \nabla_{a_s} \caFRT \|_0.
\end{equation}
The vectors $a_s \in \mathbb{Z}^2 \setminus \{0\}$
 belong to a finite difference system 
$\mathscr{N}$ with $S \geq 2$ elements. 
For $a \in \mathbb{Z}^2,$ we let
\begin{equation}\label{eq:pottsPriorDiscretization}
\| \nabla_{a} \caFRT \|_0 = | \{ i = (i_1, i_2) : |\caFRT_{i, :} - \caFRT_{i + a, :}|_2 \neq 0 \}|.
\end{equation}
where we use the notation $\caFRT_{i, :} = (\caFRT_{i,1}, ..., \caFRT_{i,K}) \in \mathbb{R}^K$
to denote the data located in the pixel with coordinates $i \in \mathbb{Z}^2.$
Further, we use the symbol $|x|_2$ to denote the Euclidean norm $|x|_2 = (\sum_j x_j^2)^{1/2}.$
Here we use an eight-connected neighborhood 
represented by the finite difference system
\begin{equation}\label{eq:N1}
	 \mathscr{N} = \{ (1,0), (0,1), (1,1), (1,-1)\}.
\end{equation}
with the weights 
$
	\omega_{1/2} = \sqrt{2} - 1$ and $\omega_{3/4} =  1 - \frac{\sqrt{2}}{2}.
$
For details, we refer to \cite{chambolle1999finite, storath2014fast}.

\subsection{Relaxation}
\label{subsec:relaxLearning}

Since solving Eq.~\eqref{eq:proposedModel} is computationally extremely hard, 
we impose the following simplifications to make it tractable:
For the feature learning part, 
we propose to replace the strict $\ell_0$ term in Eq.~\eqref{eq:pottsPriorDiscretization} by the smooth non-convex sparsity promoting surrogate function
\begin{equation}\label{eq:l0_log_approx}
    \| \nabla_{a} \Tensor{V} \|_{0,\nu} = \sum_{i} \log (1 + \nu |\caFRT_{i, :} - \caFRT_{i + a, :}|_2^2),
\end{equation}
which is a good approximation of the jump penalty with equality in the limit of its parameter $\nu$, cf.  \cite{kiechle2015}. Further, we let $\varepsilon = 0$ in Eq.~\eqref{eq:proposedModel}
which leads to minimizing the (preliminary) cost function
\begin{equation} \label{eq:sparsity_objective}
f(\AOF{}) = \sum_{ s=1}^S  \omega_s \| \nabla_{a_s} \sigma(\AOF{} \SCI) \|_{0, \nu}
\end{equation}
for learning the filters.
We note that the latter assumption frees us from performing segmentation during learning and thus allows us to proceed sequentially instead of in an alternating way. Further, the relaxation of the jump penalty imposes less penalty for small variations in the data which is due to the absence of regularization and favorable compared with the jump term here.

The non-linear transformation  $\sigma$ of the filter responses 
in Eq.~\eqref{eq:sparsity_objective} is realized via 
 $\sigma(x):=\log(1+\mu x^2)$ with parameter $\mu >0$. Note, that $\sigma$ is smooth and symmetric,  
and that it allows to 
attenuate outliers in the filter responses.

\subsection{Learning from patch samples}
\label{subsec:learnignFromPatch}

We choose the training samples as a subset of image locations (and not all patches given by the image).
This can be motivated as follows: first, when learning convolutional filters by minimizing Eq.~\eqref{eq:sparsity_objective} we evaluate the inner products of each filter kernel $\AOF{k}$ with the pixel neighborhood at all image locations $(i,j)$ and sum them up w.r.t.\ $i,j$.
Due to overlap, calculating the whole sum results in redundant computations.
Secondly, since the data of interest consists of texture segments,
we expect repeating patterns which in turn makes the full patch set look even more redundant. 
Based on this intuition, a randomly sampled subset of patches 
should suffice to learn the features from a texture image. 
Hence,
 we only consider a fixed number $M$ of randomly sampled patches as training set.

\begin{figure}
    \includegraphics[width=\linewidth]{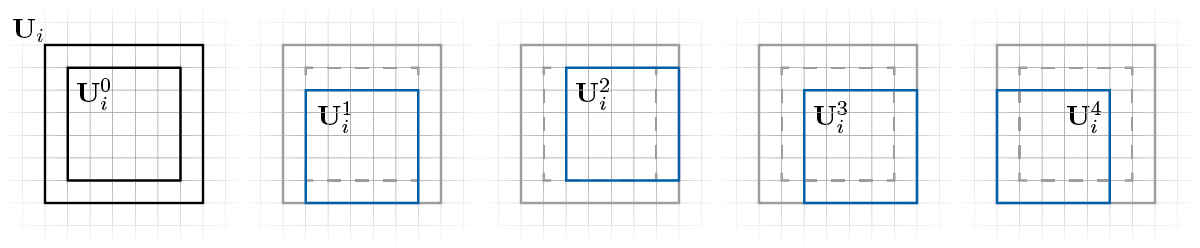}
    \caption{Illustration of an extracted super patch $\Matrix{U}_i$  and its neighboring patches $\Matrix{U}_i^{a_s}$
    with respect to the utilized finite difference system $\mathscr{N}.$}
    \label{fig:patch_crops}
\end{figure}

Formally, we modify the data in the objective function Eq.~\eqref{eq:sparsity_objective} from the jump set over features of the entire image to the empirical mean of a set of randomly sampled super-patches $\Matrix{U}_i$ and we obtain
\begin{align} \label{eq:sparsity_objective_on_patches}
f(\AOF{}) &= \frac{1}{M} \sum_i^M \sum_{s = 1}^S \omega_{a_s} \Vert \nabla_{a_s} \sigma ( \AOF{} \Matrix{U}_i) \Vert_{0, \nu} .
\end{align}
Here, the super-patches' support templates are extensions of the $\sqrt{n} \times \sqrt{n}$ support template of the filters which additionally take the considered finite difference stencil into account.
We refer to Figure~\ref{fig:patch_crops} for a detailed visualization.
For first order finite differences, a corresponding one pixel neighborhood of the considered $\sqrt{n} \times \sqrt{n}$ template is sufficient.
We then generate different crops from these super-patches according to the direction of the finite difference discretization and evaluate the inner products w.r.t.\ these crops (and apply $\sigma$).
Finally, we apply the respective finite difference operator to the obtained result.  

\subsection{Constraints}

\label{subsec:constraints}

In order to avoid trivial solutions such as the zero kernel and redundancies, 
we impose several constraints on the filters.

\subsubsection{Norm and coherence constraints} \label{sec:norm_coherence_constraints}

Following \cite{hawe2013}, we impose norm and coherence constraints. 
To prevent the filter coefficients from shrinking to zero,
we require the Euclidean norm of each filter to equal one, i.e.,
\begin{equation} \label{eq:filter_norm_constraint}
\Vert \AOF{k} \Vert_2 = \sqrt{\sum_{i=1}^n (\AOF{k})_{i}^2} = 1, \qquad k=1,\dots,K.
\end{equation}
Here, $n$ is the number of coefficients in a single filter. For brevity,
we consider 2D filters of quadratic support with size $\sqrt{n}\times\sqrt{n}.$ 
The extension to filters supported on a rectangle is obvious.
Geometrically, the norm constraint implies that each filter is an element of  the $(n-1)$-dimensional sphere $\mathcal{S}_{n-1}$ in $\R{n}$, and that the filter set constitutes a product of $K$ such spheres. This structure is commonly referred to as oblique manifold, i.e.\ matrices in $\R{n \times K}$ with normalized columns, denoted by
\[
\AOF{}^{\top} \in \mathcal{S}_{n-1}^{\times K}.
\]
In addition, we use the coherence penalty, cf. \cite{hawe2013},
\begin{equation} \label{eq:incoherence_penalty}
r(\AOF{}) = - \sum_{1\leq i \leq j \leq K} \log (1 - \langle\AOF{i},\AOF{j}\rangle^2 )
\end{equation}
to well separate these vectors on the sphere.
In particular, this soft constraint avoids pairwise collinear filters. 
We note that a minimum of that function is clearly achieved if the filters are orthogonal to each other, i.e., if the filter set lies in the corresponding Stiefel manifold. However, in the context of sparse coding, imposing such orthogonality directly as a hard constraint has turned out to be too restrictive, see for instance\ \cite{hawe2013,kiechle2015}.

\subsubsection{Zero-mean constraint}

The mean over the patch is a distinguished feature with special discriminative power.
We consider it a seeded filter in the filter bank and learn the other filters in its orthogonal complement. This means that we learn filters with vanishing first order moments,
i.e., filters whose coefficients sum up to zero, 
\begin{equation} \label{eq:filter_zeromean_constraint}
\sum_{i=1}^n \AOF{k,i} = 0.
\end{equation}
We note that these filters do not see the patch mean
which might vary, for instance, due to small differences in lighting or contrast.
Geometrically, the filters that satisfy Eq.~\eqref{eq:filter_zeromean_constraint}
are contained in the hyperplane which contains the origin and which is orthogonal to 
 $\ones_n = (1,\ldots,1)$. 
Hence,  the set of feasible solutions is a Riemannian manifold as well \cite{kiechle2015}. We denote it by
\begin{equation}
\label{eq:DefManifold}
\mathcal{R} = \left( \mathcal{S}_{n-1} \cap \ones_n^{\perp} \right)^{\times K}
\end{equation}
in the following. The Riemannian structure is important for the optimization procedure used later on.

\subsubsection{Central moment constraint}

\begin{figure}
    \centering
    \begin{subfigure}[b]{0.48\linewidth}
        \includegraphics[width=\textwidth]{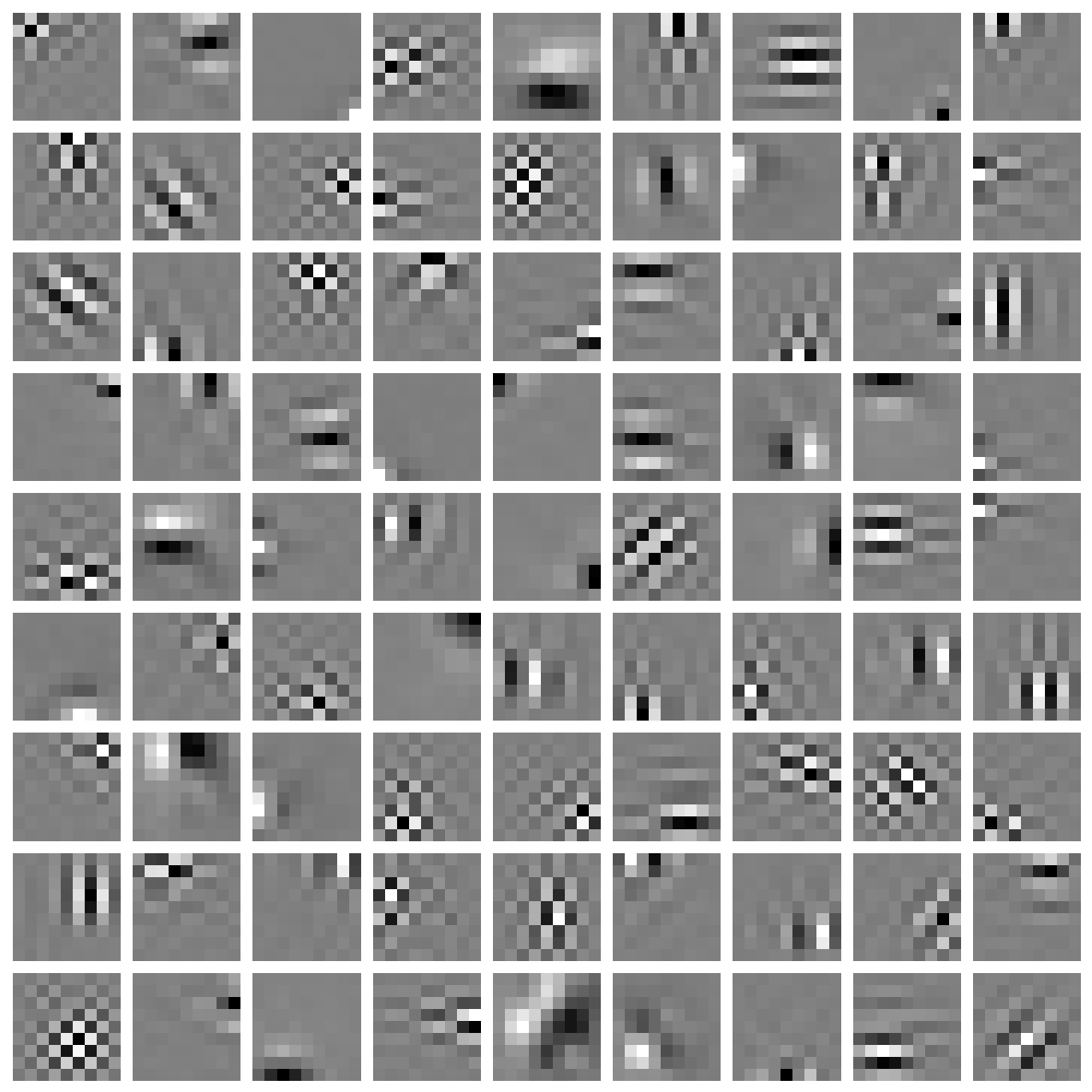}
        \caption{}
        \label{fig:uncentered_filter_set}
    \end{subfigure}
    ~
    \begin{subfigure}[b]{0.48\linewidth}
        \includegraphics[width=\textwidth]{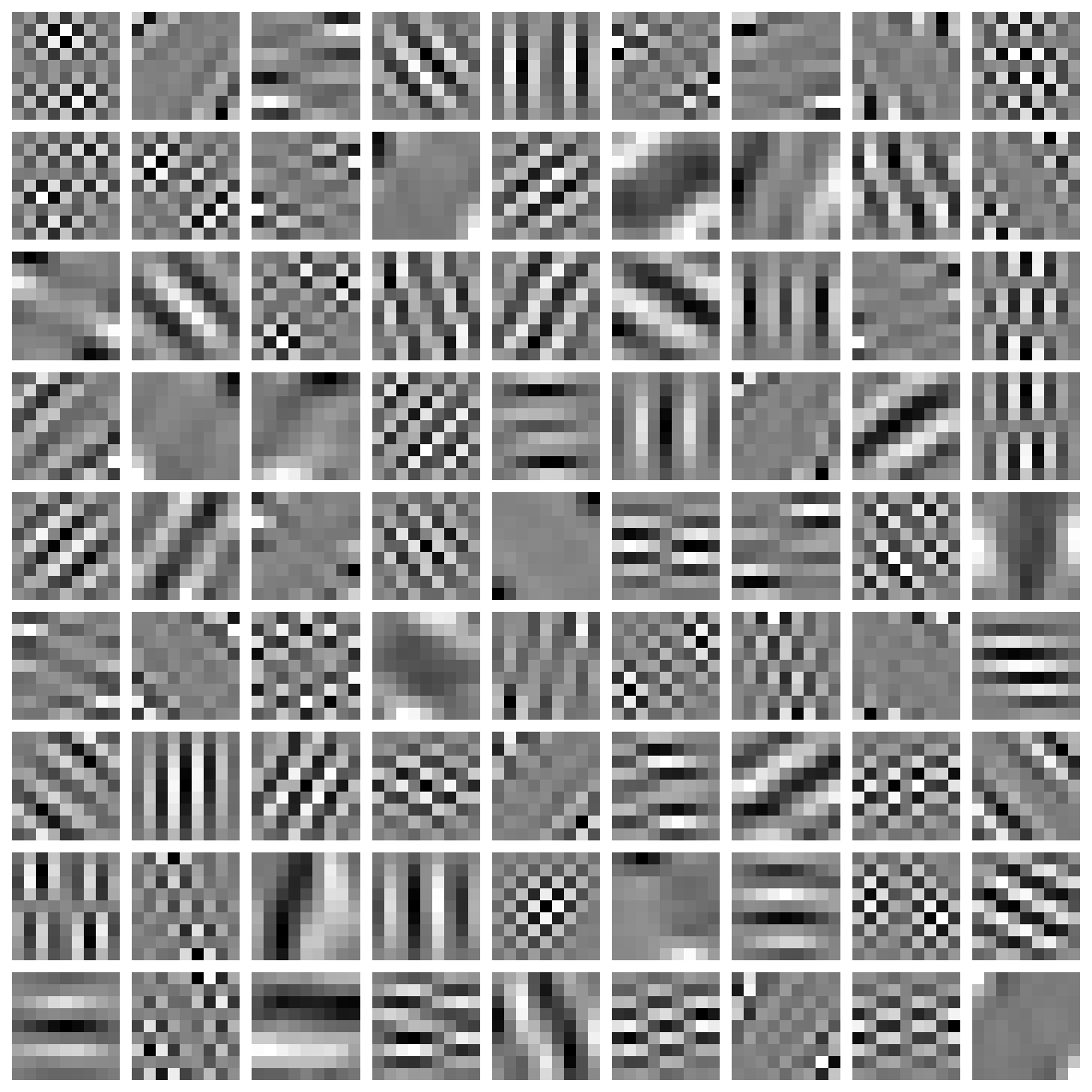}
        \caption{}
    \end{subfigure}
    \caption{
    Effect of the proposed central moment constraint.
    Two sets of filters learned from the same gray-scale cartoon image. 
    (Dark and light pixels represent negative and positive filter coefficients respectively, while neutral grey indicates coefficients equal or close to zero.)
    The filters in~\emph{(a)} were learned with the coherence constraint from Eq.~\eqref{eq:incoherence_penalty} but
    without the centroid constraint in Eq.~\eqref{eq:centered_moment_penalty}. 
    In contrast, the filters in~\emph{(b)} were learned using both 
    the coherence constraint Eq.~\eqref{eq:incoherence_penalty}
    and the central moment constraint Eq.~\eqref{eq:centered_moment_penalty}. 
        It is clearly visible
that the effective support sizes of many filters in~\emph{(a)} are in fact much smaller than $9 \times 9,$
and that some shifted versions of the same filter can be identified among all filters.
These undesirable effects are significantly reduced in~\emph{(b)}.
}
\label{fig:uncetered_and_centralized_filters}
\end{figure}

It might happen that there are two minimizers of  Eq.~\eqref{eq:sparsity_objective}, which adhere to norm and coherence constraints, and which are shifted versions of each other;
see Figure~\ref{fig:uncentered_filter_set}.
Also note that, there, the effective support size of many filters is much smaller than the prescribed maximum $9 \times 9$ filter size.

To avoid learning shifted versions of the same filter, we propose a constraint on the centroid of the squared filter coefficients. 
Intuitively speaking, by penalizing off-centered centroids 
of the pointwise squared (real-valued) filters, 
we prevent learning filters that are shifted versions of their centered twin.
To be more precise, we consider a filter $\AOF{k}$ and notice that, by the employed normalization, we have 
$\hat{\boldsymbol{\Phi}}_{k}^{\top}\hat{\boldsymbol{\Phi}}_{k}$ 
$= \sum_i \sum_j (\AOF{k})_{ij}^2 = 1$ where
$\hat{\boldsymbol{\Phi}}_{k}$ denotes the vectorized 2D filter $\AOF{k}.$
Thus, the pointwise square ${\boldsymbol{\Psi}}_{k}$ defined 
by $(\boldsymbol{\Psi}_{k})_{ij} = (\AOF{k})_{ij}^2$ can be viewed as a discrete 2D probability distribution. Hence, we may compute the components of the center of mass of this distribution by 
\begin{equation}
\bar{c}_{k,x} = \hat{\boldsymbol{\Phi}}_{k}^{\top} \Matrix{P}_x \hat{\boldsymbol{\Phi}}_{k},
\quad
\bar{c}_{k,y} = \hat{\boldsymbol{\Phi}}_{k}^{\top} \Matrix{P}_y \hat{\boldsymbol{\Phi}}_{k}.
\end{equation}
Here $\Matrix{P}_x$ is a diagonal matrix realizing the first moment with respect to the $x$-direction
$\sum_{ij} i ({\boldsymbol{\Psi}}_{k})_{ij},$
and $\Matrix{P}_y$ is given analogously.
\begin{figure}[htbp]
    \centering
    \begin{subfigure}[b]{0.48\linewidth}
        \includegraphics[width=\textwidth]{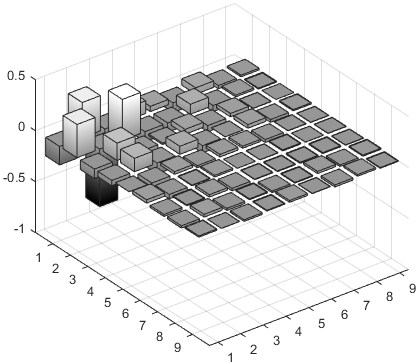}
        \caption{}
                \label{fig:filter_centroid_a}
    \end{subfigure}
    ~
    \begin{subfigure}[b]{0.48\linewidth}
        \includegraphics[width=\textwidth]{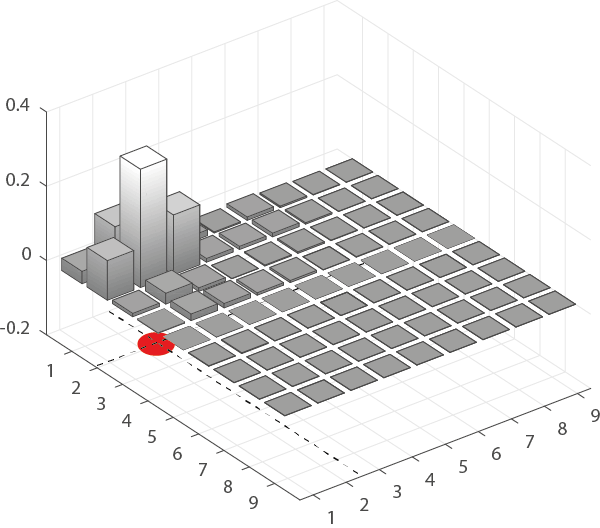}
                \caption{}
        \label{fig:filter_centroid_b}
    \end{subfigure}
    \caption{The coefficients $(\boldsymbol{\Phi}_1)_{ij}$ of the first filter $\boldsymbol{\Phi}_1$ from the learned set $\boldsymbol{\Phi}$ depicted in Figure~\ref{fig:uncentered_filter_set} without the centroid constraint~(a) and its mass distribution~(b). The red circle denotes the centroid $(c_{1,x}, c_{1,y})$.}
    \label{fig:filter_centroid}
\end{figure}
We further employ the normalization 
\(
c_{k,x} = (\bar{c}_{k,x} - \tfrac{\sqrt{n}+1}{2})/\tfrac{\sqrt{n}-1}{2})
\)
and the analogous normalization for $c_{k,y}$  
to obtain quantities $c_{k,x}, c_{k,y}$  centered at $0$ with range between $-1$ and $1.$ 
For a filter centered around the origin,
we require $c_{k,x}, c_{k,y}$ to be close to zero.
To this end, we here use the (convex) penalty
\begin{align} \label{eq:centered_moment_penalty}
h(\AOF{k}) &= \sum_{k=1}^K -\log [ (1-c_{k,x}^2) (1 - c_{k,y}^2) ] + \frac{1}{2} (c_{k,x}-c_{k,y})^2.          \end{align}
The effects of the central moment conditions
are illustrated in Figure~\ref{fig:uncetered_and_centralized_filters}
and in Figure~\ref{fig:filter_centroid}.

\subsection{Simplified learning problem and numerical optimization}
\label{subsec:SimplifiedlearningOpt}

Summing up the considerations of this section, we propose the relaxed objective Eq.~\eqref{eq:sparsity_objective_on_patches} with the soft coherence constraint Eq.~\eqref{eq:incoherence_penalty} and the soft shift constraint Eq.~\eqref{eq:centered_moment_penalty} 
which reads 
\begin{equation} \label{eq:learning_function}
E(\AOF{}) = f(\AOF{}) + \lambda r(\AOF{}) + \kappa h(\AOF{}).
\end{equation}
Here $\lambda, \kappa$ are positive parameters.
The learning objective Eq.~\eqref{eq:learning_function} is a smooth non-convex function.
The hard constraints (norm constraints, vanishing first moments) are encoded in the manifold 
$\mathcal{R}$ defined by Eq.~\eqref{eq:DefManifold}.
Equipped with this notation, the learning task reads
\begin{equation} \label{eq:final_learning_problem}
\AOF{}^{\star} \in \arg \min_{\AOF{}^{\top} \in \mathcal{R}}  E(\AOF{}).
\end{equation}

\begin{figure}    \centering
    \begin{subfigure}[b]{0.325\linewidth}
        \includegraphics[width=\textwidth]{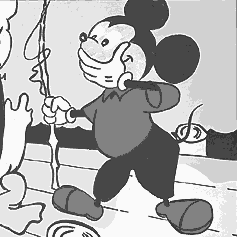}
    \end{subfigure}
    \begin{subfigure}[b]{0.325\linewidth}
        \includegraphics[width=\textwidth]{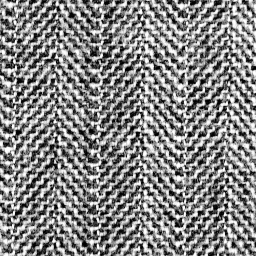}
    \end{subfigure}
    \begin{subfigure}[b]{0.325\linewidth}
        \includegraphics[width=\textwidth]{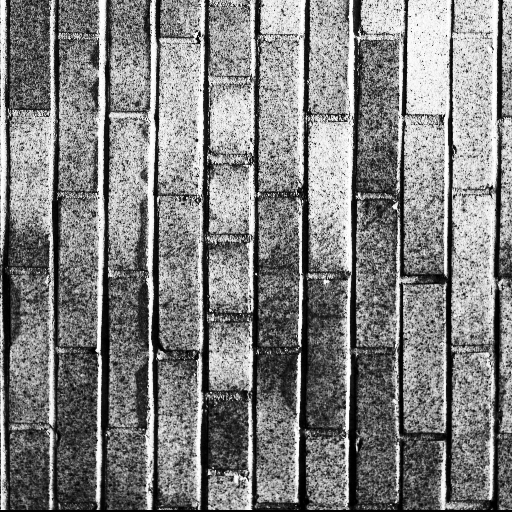}
    \end{subfigure}
    \\
    \vspace{1mm}
    \begin{subfigure}[b]{0.325\linewidth}
        \includegraphics[width=\textwidth]{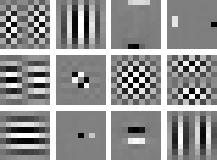}
        \caption{cartoon image}
    \end{subfigure}
    \begin{subfigure}[b]{0.325\linewidth}
        \includegraphics[width=\textwidth]{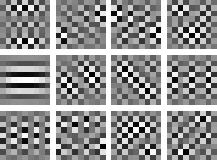}
        \caption{Brodatz 4}
    \end{subfigure}
    \begin{subfigure}[b]{0.325\linewidth}
        \includegraphics[width=\textwidth]{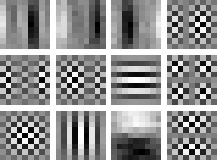}
        \caption{Brodatz 12}
    \end{subfigure}
    \caption{Filter sets (bottom) learned from different input images (top). \Idea{referenz zum Mickey Bild und zu Brodatz}}
    \label{fig:learned_filters_from_various_images}
\end{figure}

In order to solve Eq.~\eqref{eq:final_learning_problem} numerically,
we apply an efficient scheme that exploits the geometric structure 
of the manifold $\mathcal{R}$ \cite{hawe2013,kiechle2015}.
For a general introduction to gradient methods on matrix manifolds, we refer to \cite{absil2009}.
The approach of  \cite{hawe2013,kiechle2015} consists of a geometric variant of the conjugate gradients method with backtracking line-search and an Armijo step-size rule.
For a detailed explanation, we refer the reader to the aforementioned references. 
In our setup, the steps of one iteration are as follows:
\begin{enumerate}
\item compute the Euclidean gradient of the learning function Eq.~\eqref{eq:learning_function} at the current estimate (the derivation of the Euclidean gradient can be found in the appendix);
\item project the Euclidean gradient onto the tangent space of the manifold at the estimate to obtain the Riemannian gradient;
\item compute the new descent direction by linear combination of the Riemannian gradient and the descent direction of the previous iteration via parallel transport;
\item perform a backtracking line search along the geodesic in the descent direction emanating from the current estimate to obtain an optimal step size using the Armijo rule. That means, we iteratively reduce the step size until the objective function decreases;
\item update the estimate.
\end{enumerate}
We start the procedure with a random initialization in $\mathcal{R}$ and iterate until the Frobenius norm of the Riemannian gradient falls below the threshold of
$10^{-5}$. For illustration purposes, Figure~\ref{fig:learned_filters_from_various_images} depicts the learned filter sets for different images.

\subsection{Extension to vector-valued images} \label{sec:vectorvalued}
\label{subsec:ExtMultiChan}

So far, we only considered gray-scale texture images whereas textured images often have multiple channels, for instance RGB color images.
We extend our method for the case when the image $\Matrix{U}$ is vector-valued with $L$ channels, that is, if $\Matrix{U}_{ij} \in \mathbb{R}^L.$
Let $\SCI_{i,l}^{a_s}$ the $i$-th patch cropped according to direction $a_s$ in channel $l \in 1,\dots,L$.
Intuitively, different channels of an image should require different filter sets such that spatial homogeneity of filter responses can be achieved.
To that end, we first extend the formulation of the patch-based filter operation
\begin{equation}
\AOF{k} \SCI_{i}^{a_s} = 
\begin{bmatrix}
    \AOF{k,1} & 0 & 0 \\
    0 & \ddots & 0 \\
    0 & 0 & \AOF{k,l}
\end{bmatrix}
\begin{bmatrix}
    \Matrix{U}_{i,1}^{a_s}\\
    \vdots\\
    \Matrix{U}_{i,l}^{a_s}
\end{bmatrix}.
\end{equation}
In this work, we consider RGB images as examples of multi-channel images. Since the red, the green and the blue channels are in general highly correlated, we assume that the patch structure within each channel will be similar and set $\AOF{k,R}=\AOF{k,G}=\AOF{k,B}=\AOF{k}$. Thus, the learned filters act on the different channels in the same way. We note that this does not hinder jumps in a single channel to be detected.

\section{Segmentation stage}\label{sec:SegmetationStage}
After relaxing the model Eq.~\eqref{eq:proposedModel} in Section~\ref{sec:filter_learning}
to determine suitable filters $\AOF{},$
we here discuss the segmentation given a set of filters.
To segment the vector-valued feature image $\sigma(\Tensor{F})$
we consider the (formal) Lagrangian version of 
the discretization of Eq.~\eqref{eq:proposedModel} for fixed $\AOF{}$ to obtain
the problem
\begin{equation}\label{eq:pottsLagrange}
\argmin_{\Tensor{V}} \; \gamma \sum_{ s=1}^S  \omega_s \| \nabla_{a_s}  \Tensor{V} \|_0 + d(\Tensor{V}, \sigma(\Tensor{F} )).
\end{equation}
Here, $\gamma > 0$ is a parameter for tuning the trade-off between 
data fitting and regularity, and $\Tensor{F}$ are the filter responses of $\Matrix{U},$
i.e., $\Tensor{F}  =  \AOF{}\Matrix{U}.$

\subsection{Filter weighting based on the Mahalanobis distance}
\label{subsec:Mahalanb}

We recall that all filters were constrained to have unit norm in the learning stage.
As a result, all filter outputs are weighted equally regardless of their discriminative power.
To account for this, we utilize a data fidelity term based on the Mahalanobis distance $d$.
Here we use the covariance matrix of all feature vectors (after applying the non-linearity). With slight abuse of notation, let
\[
\Matrix{\Sigma} = \textrm{cov}(\Tensor{G}) 
\]
the $K\times K$ covariance matrix of all feature vectors in $\Tensor{G} = \sigma(\Tensor{F})$. 
To define the corresponding
Mahalanobis distance, 
we write $\Matrix{\Sigma}\Tensor{G}$ for the action of a $K\times K$ matrix  $\Matrix{\Sigma}$
on the third index of the feature image $\Tensor{G},$ 
i.e.,
$(\Matrix{\Sigma}\Tensor{G})_{ijk} = (\Matrix{\Sigma}\Tensor{G}_{ij})_k.$
Then, the Mahalanobis data fidelity reads
\begin{equation}\label{eq:pottsMahalanobis}
    \begin{split}
d(\Tensor{V}, \Tensor{G} ) 
     &=
\sum_{ij} | ({\Matrix
     {\Sigma}^{-1/2}} (\Tensor{V}_{ij} - \Tensor{G}_{ij})|^2_2 \\
     &= 
    \| {\Matrix
     {\Sigma}^{-1/2}} (\Tensor{V} - \Tensor{G})\|^2_2.  
     \end{split}
\end{equation}
We observed that the results slightly improve 
 when we normalize $\Matrix{\Sigma}^{-1/2}$ by $\max_{ij} \, (\Matrix{\Sigma}^{-1/2})_{ij}.$

\subsection{Variational partitioning of the feature images}
\label{ssec:plainPottsSegmentation}

By the previous considerations in Section~\ref{subsec:Mahalanb}, 
we have to solve the minimization problem Eq.~\eqref{eq:pottsLagrange}
with the Mahalanobis data term.
To that end, we plug $\Tensor{V} = \Matrix{\Sigma}^{1/2} \Tensor{U}$
into Eq.~\eqref{eq:pottsLagrange}
to obtain the problem
\begin{equation} \label{eq:pottsMalTrans}
\argmin_{\Tensor{U}} \; \gamma \sum_{ s=1}^S  \omega_s \| \nabla_{a_s} \Matrix{\Sigma}^{1/2} \Tensor{U} \|_0 + \Vert \Tensor{U} -  \Matrix{\Sigma}^{-1/2} \Tensor{G} \Vert^2_2.
\end{equation}
We observe that the $\ell_0$ prior is invariant to invertible matrices acting in the third dimension, i.e.,
$\| \nabla_{a_s} \Matrix{\Sigma}^{1/2} \Tensor{U} \|_0 = \| \nabla_{a_s} \Tensor{U} \|_0.$ 
Therefore, the problem Eq.~\eqref{eq:pottsMalTrans} is equivalent to the problem
\begin{align}\label{eq:pottsMahalanobisPlain}
\Tensor{U}^* = \argmin_{\Tensor{U}} \; \gamma \sum_{s=1}^S  \omega_s \| \nabla_{a_s} \Tensor{U} \|_0 + \Vert \Tensor{U} -  \Matrix{\Sigma}^{-1/2} \Tensor{G}  \Vert^2_2.
\end{align}
We observe that this constitutes
a classical (vector-valued) piecewise constant Mumford-Shah problem 
for data
$\Matrix{\Sigma}^{-1/2} \Tensor{G}$ with an $\ell_2$ norm data term.
This  is a challenging optimization problem in its own,
but there are well-working approximate strategies available.
Here, we utilize the ADMM-based method developed in \cite{storath2014jump, storath2014fast}.
Although computationally more demanding than other recent approaches \cite{xu2011image, cheng2014feature, nguyen2015fast}, this method currently gives
 the best quality in practice; see the comparison in \cite{nguyen2015fast}.

\subsection{Obtaining the label map}
\begin{figure}[t]
    \centering
    \begin{subfigure}[b]{0.48\linewidth}
        \includegraphics[width=\textwidth]{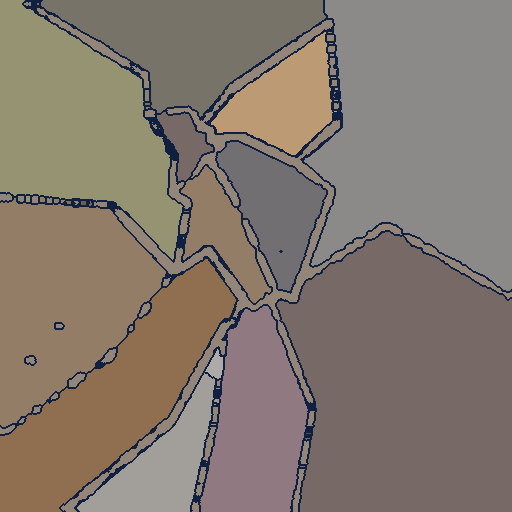}
        \caption{}
    \end{subfigure}
    ~
    \begin{subfigure}[b]{0.48\linewidth}
        \includegraphics[width=\textwidth]{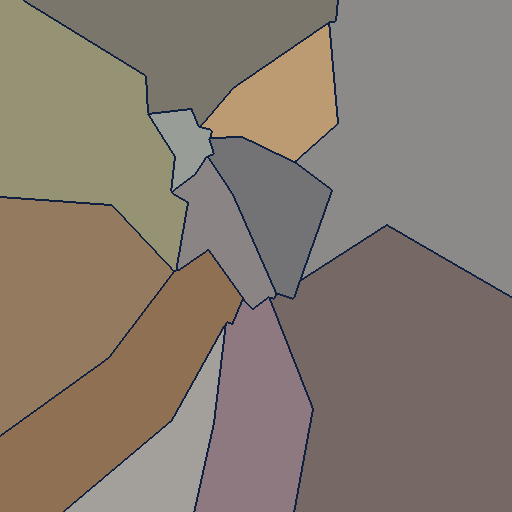}
        \caption{}
    \end{subfigure}
    \caption{In a postprocessing step, small spurious segments are merged into their neighboring segments. (a) Raw segmentation; (b) final segmentation after region merging.}
    \label{fig:region_merging}
\end{figure}
The result obtained from treating problem  Eq.~\eqref{eq:pottsMahalanobis} is a vector-valued piecewise constant function.
To obtain the final label map (scalar field), 
we simply utilize the sum of the vector in a pixel as (real-valued) index for a segment,
i.e., we sum up the coefficients along the feature vector at every pixel location.
We observe that segment boundaries often lead to high filter responses
which results in small spurious segments at the boundaries.
To remove these, 
we adopt the simple post-processing step from \cite{yuan2015factorization}, where small regions are merged with neighbors based on their boundary ratios.
Figure~\ref{fig:region_merging} depicts the final segmentation before and after the boundary refinement step.

\section{Experimental Results}
\begin{table*}[!htbp]
\centering
\caption{Results on the Prague Color Texture Dataset (ICPR2014 Contest).
Each row corresponds to a  segmentation quality metric,
and the arrow indicates if high or low values are better.
The first rank is marked by boldface, 
the second rank is marked by an asterisk.
\Idea{Stand 2016-06-03}}
\label{tbl:PragueColor_ICPR2014}
\begin{tabular}{@{} l *{11}{S[table-format=2.3,detect-inline-weight=math,table-align-text-post=false]} @{}}\toprule
{Method}         & {TS}  & {SWA}         & {GMRF}& {AR3D} & {TFR}         & {TFR+}& {RS}           & {FSEG}& {PMCFA}            & {PCA-MS}          & {Proposed}  \\
\midrule
$\uparrow$ CS    & 59.13 & 27.06         & 31.93 & 37.24  & 46.13         & 51.25 & 46.02          & 69.02 & 75.32*             & 72.27             & \bfseries 77.73   \\
$\downarrow$ OS  & 10.89 & 50.21         & 53.27 & 59.53  & \bfseries 2.37 & 5.84* & 13.96         & 17.30 & 11.95              & 18.33             & 15.92             \\
$\downarrow$ US  & 18.79 & \bfseries 4.53 & 11.24 & 8.86   & 23.99         & 7.16  & 30.01         & 11.85 & 9.65               & 9.41              & 6.31*             \\
$\downarrow$ ME  & 10.45 & 25.76         & 14.97 & 12.54  & 26.70         & 31.64 & 12.01          & 6.28  & 4.57               & 4.19*             & \bfseries 3.93    \\
$\downarrow$ NE  & 9.93  & 27.50         & 16.91 & 13.14  & 25.23         & 31.38 & 11.77          & 5.66  & 4.63               & \bfseries 3.92    & \bfseries 3.92    \\
$\downarrow$ O   &       & 33.01         & 36.49 & 35.19  & 27.00         & 23.60 & 35.11          & 10.79 & \bfseries 4.51     & 7.25*             & 7.68              \\
$\downarrow$ C   &       & 85.19         & 12.18 & 11.85  & 26.47         & 22.42 & 29.91          & 13.75 & 8.87*              & \bfseries 6.44    & 24.24             \\
$\uparrow$ CA    &       & 54.84         & 57.91 & 59.46  & 61.32         & 67.45 & 58.75          & 77.50 & \bfseries 83.50    & 81.13             & 82.80*            \\
$\uparrow$ CO    &       & 60.67         & 63.51 & 64.81  & 73.00         & 76.40 & 68.89          & 84.11 & \bfseries 88.16    & 85.96             & 86.89*            \\
$\uparrow$ CC    &       & 88.17         & 89.26 & 91.79* & 68.91         & 81.12 & 69.30          & 86.89 & 90.73              & 91.24             & \bfseries 93.65   \\
$\downarrow$ I.  &       & 39.33         & 36.49 & 35.19  & 27.00         & 23.60 & 31.11          & 15.89 & \bfseries 11.84    & 14.04             & 13.11*            \\
$\downarrow$ II. &       & 2.11          & 3.14  & 3.39   & 8.56          & 4.09  & 8.63           & 2.60  & \bfseries 1.47     & 1.59              & 1.50*             \\
$\uparrow$ EA    &       & 66.94         & 68.41 & 69.60  & 68.62         & 75.80 & 65.87          & 83.99 & \bfseries 88.10    & 87.08             & 88.03*            \\
$\uparrow$ MS    &       & 53.71         & 57.42 & 58.89  & 59.76         & 65.19 & 55.52          & 78.25 & \bfseries 83.98    & 81.84             & \bfseries 83.98   \\
$\downarrow$ RM  &       & 6.11          & 4.56  & 4.66   & 7.57          & 6.87  & 10.96          & 4.51  & 3.76*              & 4.45              & \bfseries 3.27    \\
$\uparrow$ CI    &       & 70.32         & 71.80 & 73.15  & 69.73         & 77.21 & 67.35          & 84.71 & 88.74*             & 87.81             & \bfseries 89.03   \\
$\downarrow$ GCE &       & 17.27         & 16.03 & 12.13  & 15.52         & 20.35 & 11.23          & 10.82 & \bfseries 6.51     & 8.33              & 7.40*             \\
$\downarrow$ LCE &       & 11.49         & 7.31  & 6.69   & 12.03         & 14.36 & 7.70           & 7.51  & \bfseries 3.92     & 5.61*             & 5.62              \\
$\downarrow$ dD  &       &               &       &        &               &       & 18.52          &       & 10.13              & 9.06*             & \bfseries 8.57    \\
$\downarrow$ dM  &       &               &       &        &               &       & 23.67          &       & 6.41               & 5.88*             & \bfseries 5.30    \\
$\downarrow$ dVI &       &               &       &        &               &       & \bfseries 13.31 &       & 15.80             & 14.54*            & 14.88             \\
\bottomrule
\end{tabular}
\end{table*}

We implemented the proposed learning and segmentation method in Matlab. For the segmentation step described in Section~\ref{ssec:plainPottsSegmentation}, we used of the toolbox Pottslab\footnote{Available at \url{http://pottslab.de}.}. In addition, we utilize the region merging implementation from \cite{yuan2015factorization} as post-processing.
The experiments were conducted on a desktop computer with an Intel i7-3930K processor with 3.2~GHz.

We compare the segmentation results produced by our method with existing algorithms on two different datasets. For a quantitative comparison, we use the well-known Prague texture segmentation dataset which comprises mosaics of color and grayscale textures. In addition, we show that the same method is also effective in segmenting the histology images from \cite{mccann2014}.

\subsection{Prague texture dataset} \label{subsec:experiments_prague}
The Prague texture segmentation dataset \cite{haindl2008texture} consists of 80 texture mosaics which are synthetically generated from random compositions of 114 different textures from 10 thematic categories. Color (RGB) and grayscale versions of this dataset are available along with the respective ground truth segment map and each texture mosaic is of size $512 \times 512$ pixels and the number of segments varies between 3 and 12.
For a quantitative comparison, we produce segmentations of the large color texture dataset -- used in the ICPR 2014 contest --
and evaluate them against their ground truth using region-based metrics correct segmentation (CS), over-segmentation (OS), under-segmentation (US), missed error (ME), noise error (NE); pixel-based metrics omission error (O), commission error (C), class accuracy (CA), recall (CO), precision (CC), type I error (I.), type II error (II.), mean class accuracy estimate (EA), mapping score (MS), root mean square proportion estimation error (RM),  comparison index (CI); and consistency-based metrics global consistency error (GCE) and local consistency error (LCE). If available, we also report the Mirkin metric (dM), Van Dongen metric (dD) as well as the variation of information (dVI). For computing these metrics, we use the benchmark provided by the authors of the Prague dataset at \cite{haindl2008online} and where a detailed definition of above metrics can be found.

For each of the 80 texture mosaics in the benchmark, we learn a separate set of filters $\AOF{}$ and compute the segmentation based on the these filter outputs subsequently. The parameters for learning the features and performing the Potts segmentation are set empirically and remain fixed for all instances in the dataset. The learned filter sets contain $K=41$ filters of size $9\times 9$ each and are learned from $M=50\,000$ patches that are drawn from the mosaic (uniform random sampling). 
In principle, the objective function Eq.~\eqref{eq:proposedModel} does not require 
the filters $\AOF{1}, \ldots, \AOF{K}$ to be of equal size.
For simplicity, we used filters of identical size, and note that filters of smaller size are included in the utilized filter set by zero-padding.
As is common practice in patch-based methods (for example \cite{nieuwenhuis2013}), we weight all pixels in the patch by a Gaussian mask to give more weight to the central pixel which leads to slightly better localized segment boundaries. In the learning problem \eqref{eq:final_learning_problem} we set the parameter of the non-linearities to $\mu = \nu = 2000$ and the weights of the coherence and moment-centering penalties to $\lambda = 10$ and $\kappa = 10$.
In the Potts segmentation that follows we require the weight that trades data fidelity against spatial homogeneity of the solution and therefore effectively influences the degree of over-segmentation. Empirically, we find $\gamma = 0.03$ to provide a good trade-off between over- and under-segmentation over all benchmark images.
The texture mosaic needed in average
35~min for the learning stage and 9~min for the segmentation stage.

\begin{figure*}
    \centering
    \begin{subfigure}[b]{0.16\textwidth}
        \includegraphics[width=\textwidth]{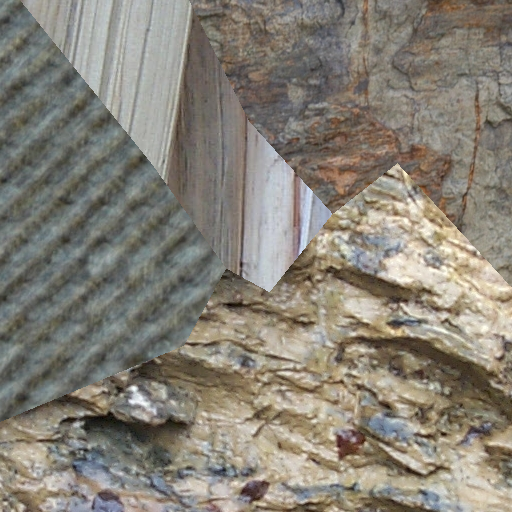}
    \end{subfigure}
    \hfill
    \begin{subfigure}[b]{0.16\textwidth}
        \includegraphics[width=\textwidth]{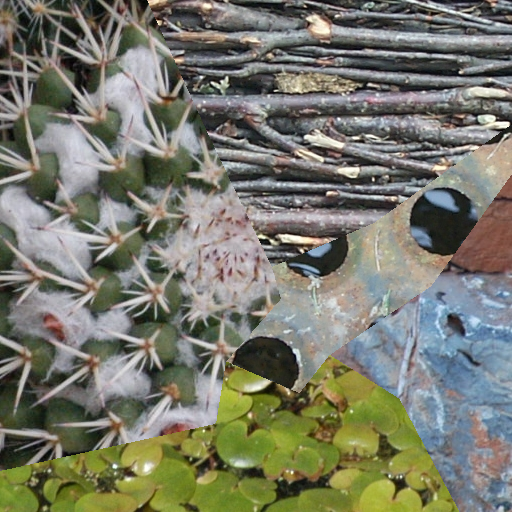}
    \end{subfigure}
    \hfill
    \begin{subfigure}[b]{0.16\textwidth}
        \includegraphics[width=\textwidth]{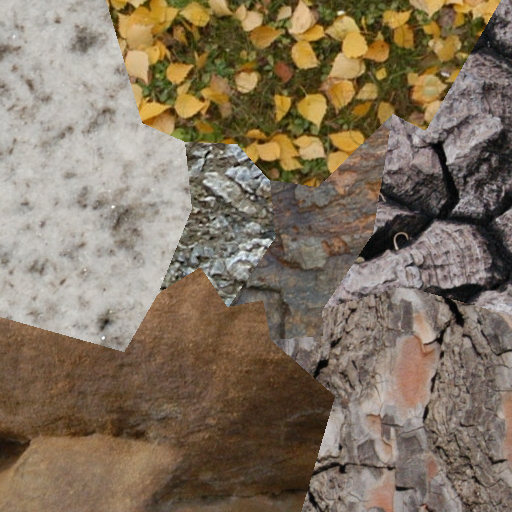}
    \end{subfigure}
    \hfill
    \begin{subfigure}[b]{0.16\textwidth}
        \includegraphics[width=\textwidth]{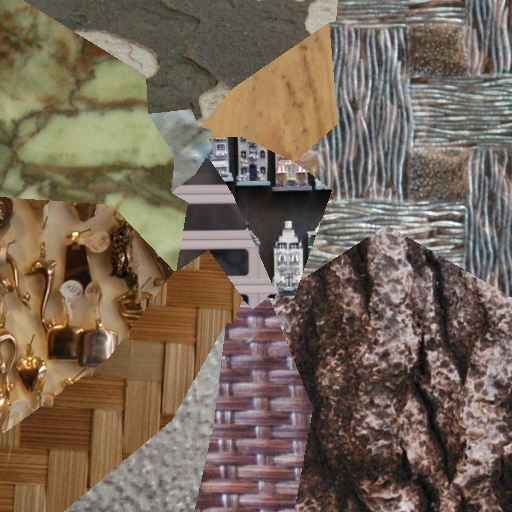}
    \end{subfigure}
    \hfill
    \begin{subfigure}[b]{0.16\textwidth}
        \includegraphics[width=\textwidth]{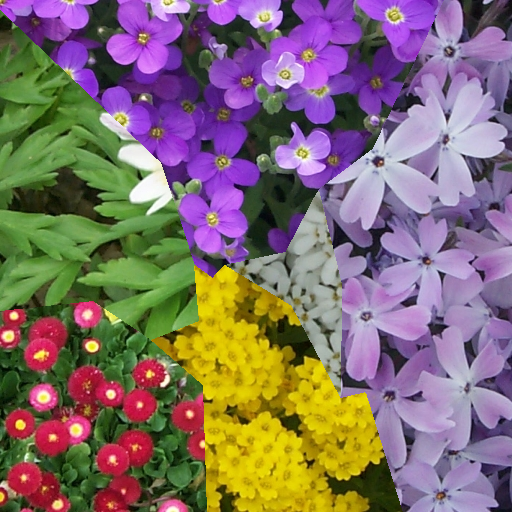}
    \end{subfigure}
    \hfill
    \begin{subfigure}[b]{0.16\textwidth}
        \includegraphics[width=\textwidth]{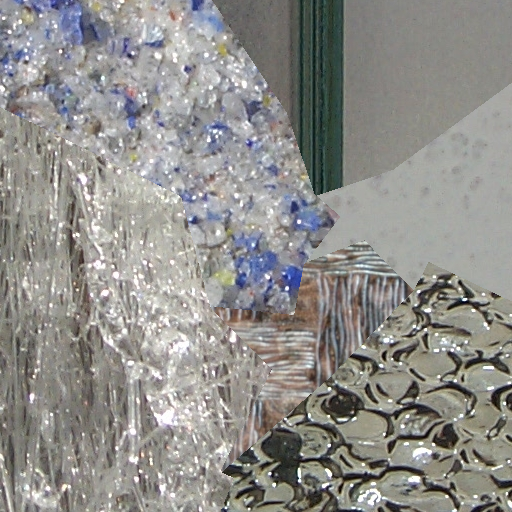}
    \end{subfigure}
    \\
    \vspace{1mm}
    \begin{subfigure}[b]{0.16\textwidth}
        \includegraphics[width=\textwidth]{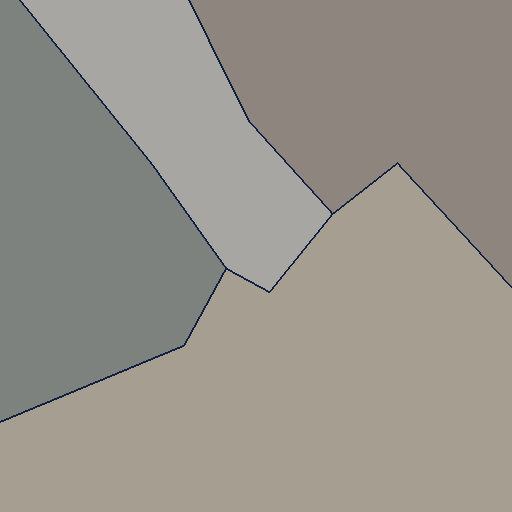}
    \end{subfigure}
    \hfill
    \begin{subfigure}[b]{0.16\textwidth}
        \includegraphics[width=\textwidth]{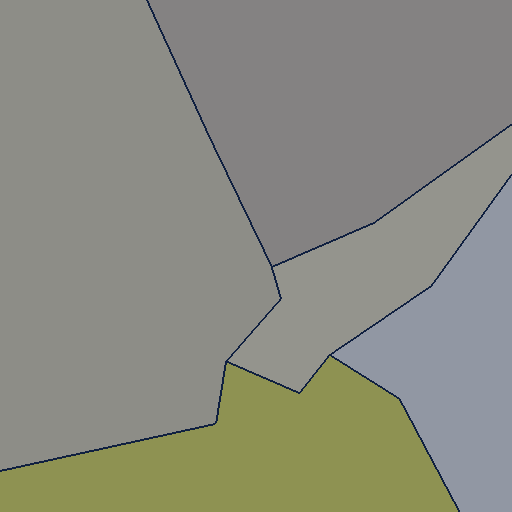}
    \end{subfigure}
    \hfill
    \begin{subfigure}[b]{0.16\textwidth}
        \includegraphics[width=\textwidth]{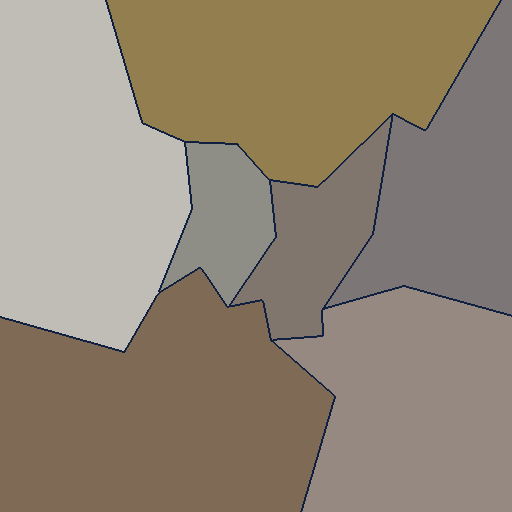}
    \end{subfigure}
    \hfill
    \begin{subfigure}[b]{0.16\textwidth}
        \includegraphics[width=\textwidth]{images/gt10_1_1.png}
    \end{subfigure}
    \hfill
    \begin{subfigure}[b]{0.16\textwidth}
        \includegraphics[width=\textwidth]{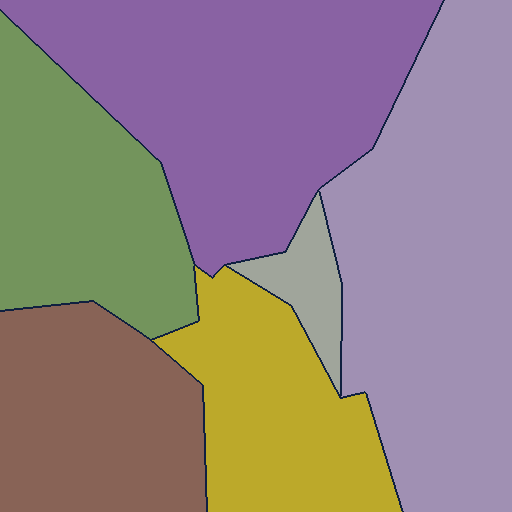}
    \end{subfigure}
    \hfill
    \begin{subfigure}[b]{0.16\textwidth}
        \includegraphics[width=\textwidth]{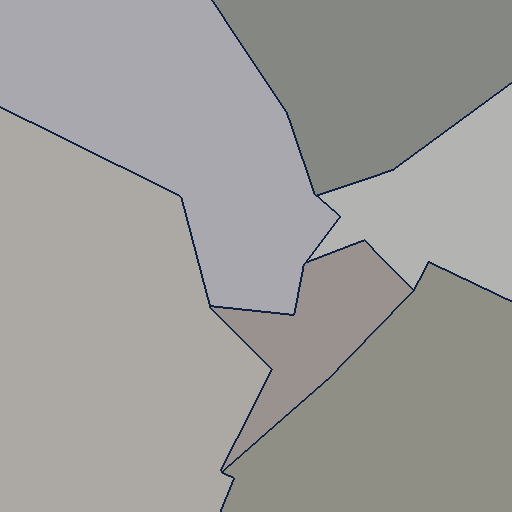}
    \end{subfigure}
    \\
    \vspace{1mm}
    \begin{subfigure}[b]{0.16\textwidth}
        \includegraphics[width=\textwidth]{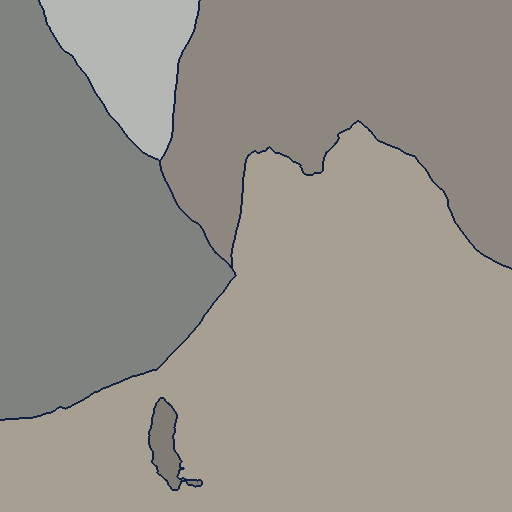}
    \end{subfigure}
    \hfill
    \begin{subfigure}[b]{0.16\textwidth}
        \includegraphics[width=\textwidth]{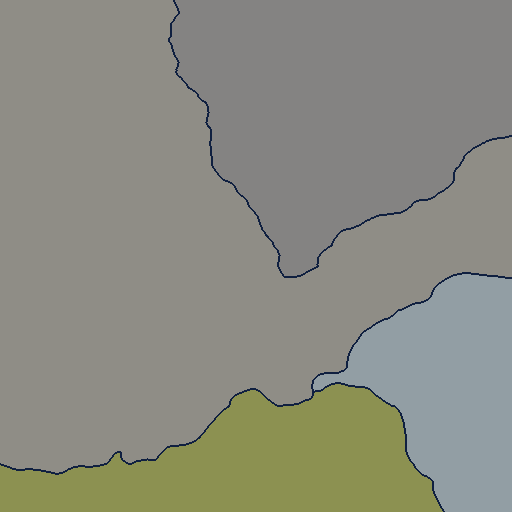}
    \end{subfigure}
    \hfill
    \begin{subfigure}[b]{0.16\textwidth}
        \includegraphics[width=\textwidth]{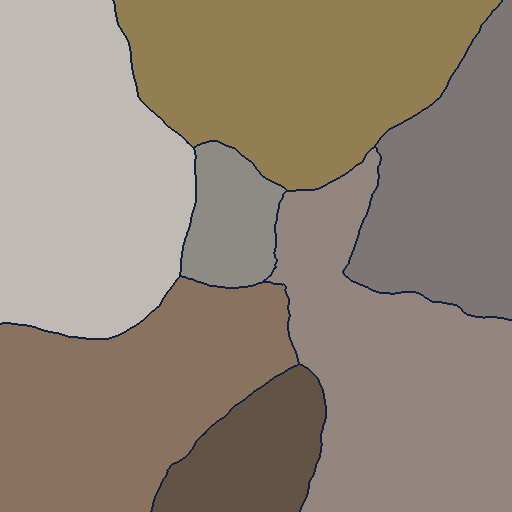}
    \end{subfigure}
    \hfill
    \begin{subfigure}[b]{0.16\textwidth}
        \includegraphics[width=\textwidth]{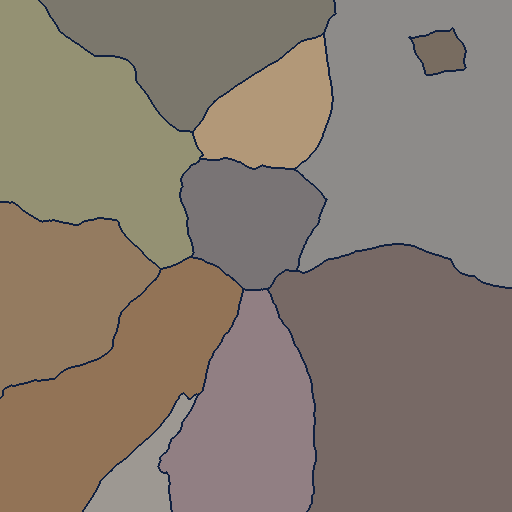}
    \end{subfigure}
    \hfill
    \begin{subfigure}[b]{0.16\textwidth}
        \includegraphics[width=\textwidth]{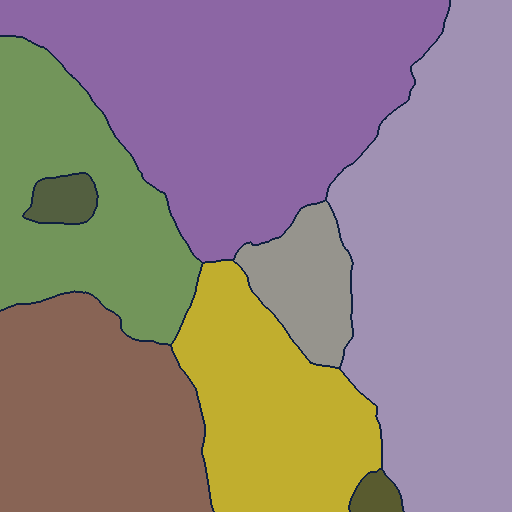}
    \end{subfigure}
    \hfill
    \begin{subfigure}[b]{0.16\textwidth}
        \includegraphics[width=\textwidth]{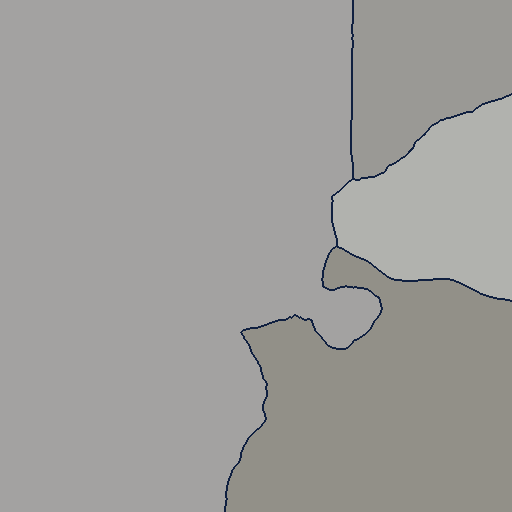}
    \end{subfigure}
    \\
    \vspace{1mm}
    \begin{subfigure}[b]{0.16\textwidth}
        \includegraphics[width=\textwidth]{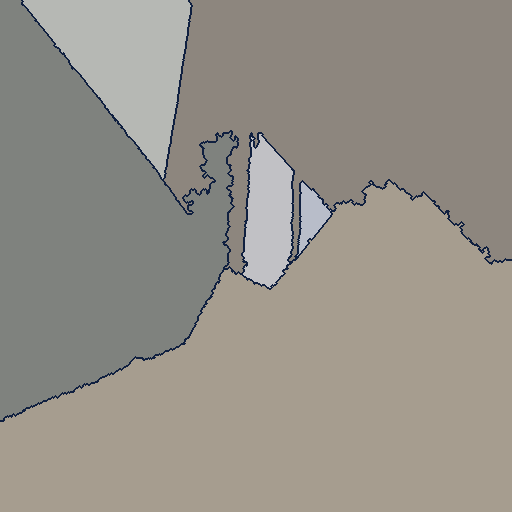}
    \end{subfigure}
    \hfill
    \begin{subfigure}[b]{0.16\textwidth}
        \includegraphics[width=\textwidth]{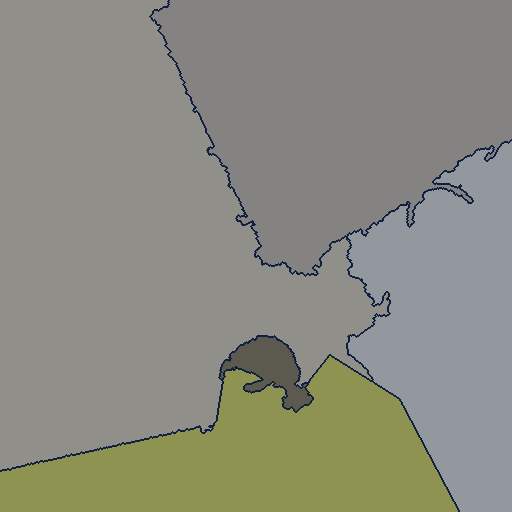}
    \end{subfigure}
    \hfill
    \begin{subfigure}[b]{0.16\textwidth}
        \includegraphics[width=\textwidth]{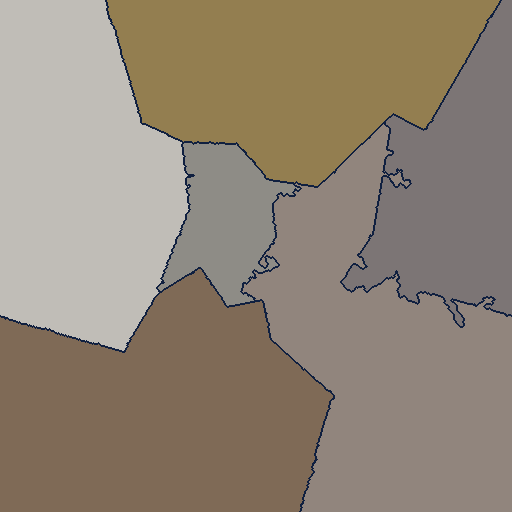}
    \end{subfigure}
    \hfill
    \begin{subfigure}[b]{0.16\textwidth}
        \includegraphics[width=\textwidth]{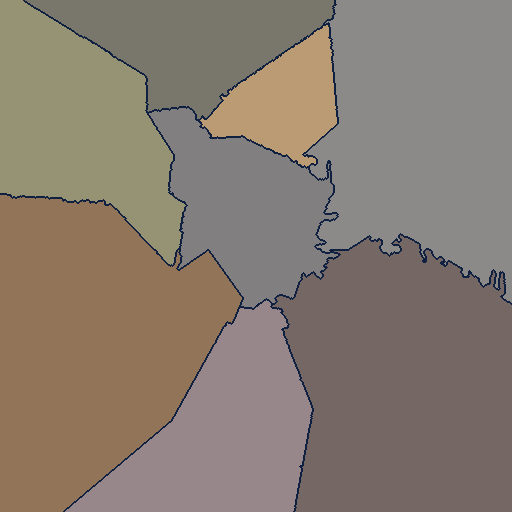}
    \end{subfigure}
    \hfill
    \begin{subfigure}[b]{0.16\textwidth}
        \includegraphics[width=\textwidth]{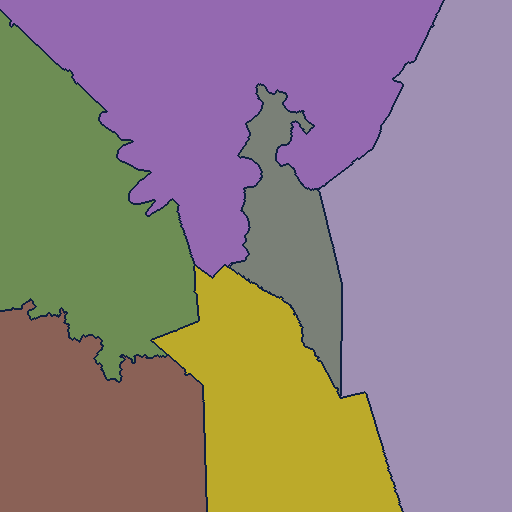}
    \end{subfigure}
    \hfill
    \begin{subfigure}[b]{0.16\textwidth}
        \includegraphics[width=\textwidth]{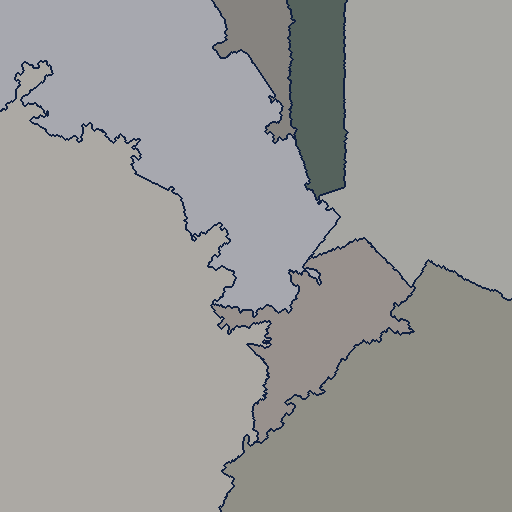}
    \end{subfigure}
    \\
    \vspace{1mm}
    \begin{subfigure}[b]{0.16\textwidth}
        \includegraphics[width=\textwidth]{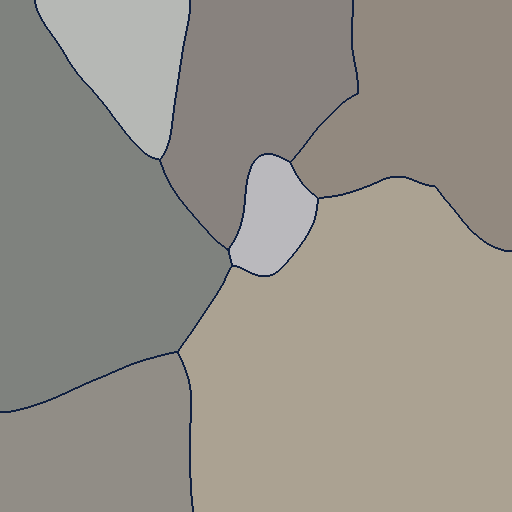}
    \end{subfigure}
    \hfill
    \begin{subfigure}[b]{0.16\textwidth}
        \includegraphics[width=\textwidth]{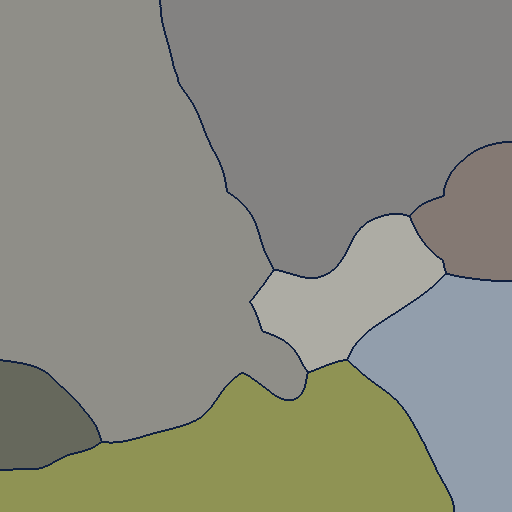}
    \end{subfigure}
    \hfill
    \begin{subfigure}[b]{0.16\textwidth}
        \includegraphics[width=\textwidth]{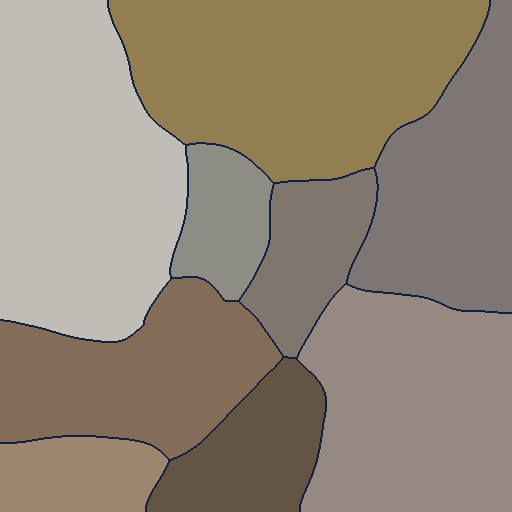}
    \end{subfigure}
    \hfill
    \begin{subfigure}[b]{0.16\textwidth}
        \includegraphics[width=\textwidth]{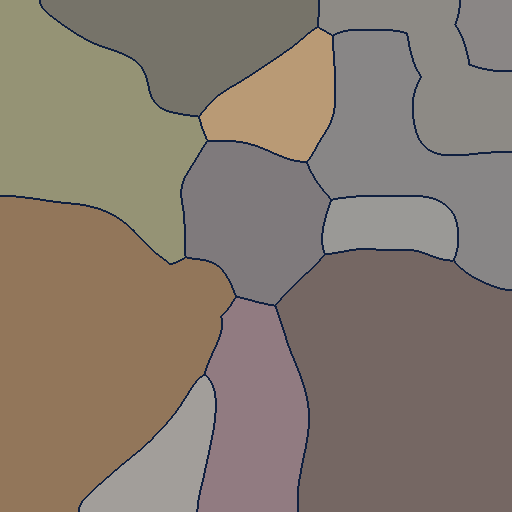}
    \end{subfigure}
    \hfill
    \begin{subfigure}[b]{0.16\textwidth}
        \includegraphics[width=\textwidth]{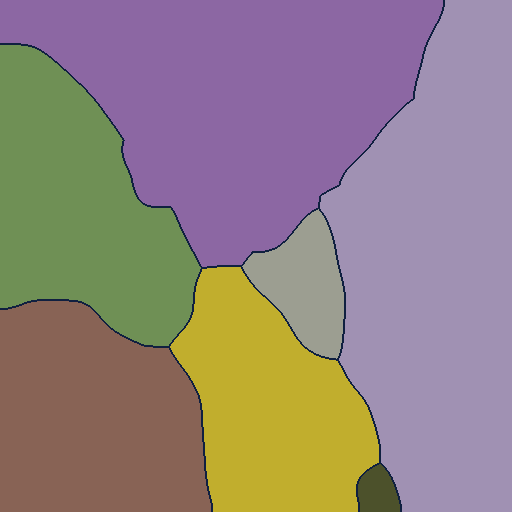}
    \end{subfigure}
    \hfill
    \begin{subfigure}[b]{0.16\textwidth}
        \includegraphics[width=\textwidth]{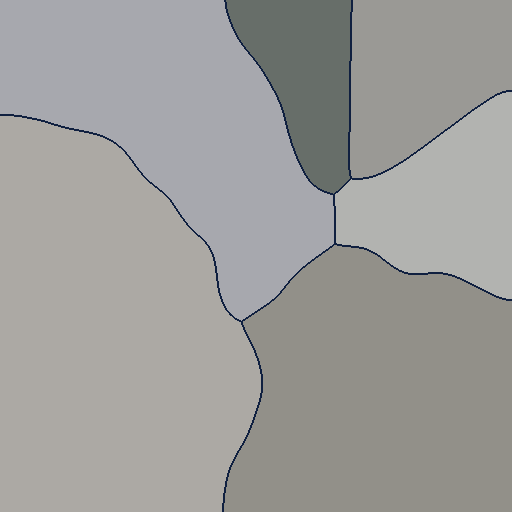}
    \end{subfigure}
    \\
    \vspace{1mm}
    \begin{subfigure}[b]{0.16\textwidth}
        \includegraphics[width=\textwidth]{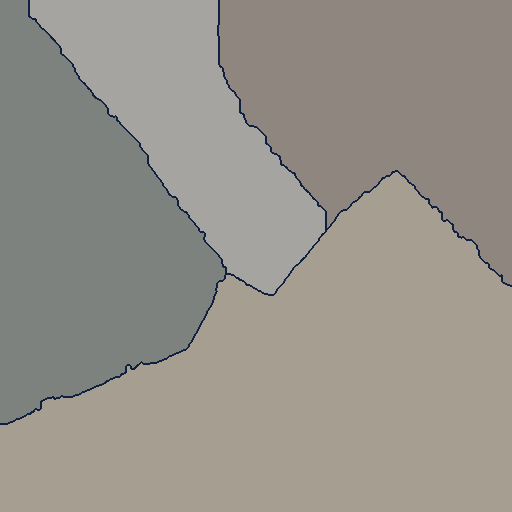}
    \end{subfigure}
    \hfill
    \begin{subfigure}[b]{0.16\textwidth}
        \includegraphics[width=\textwidth]{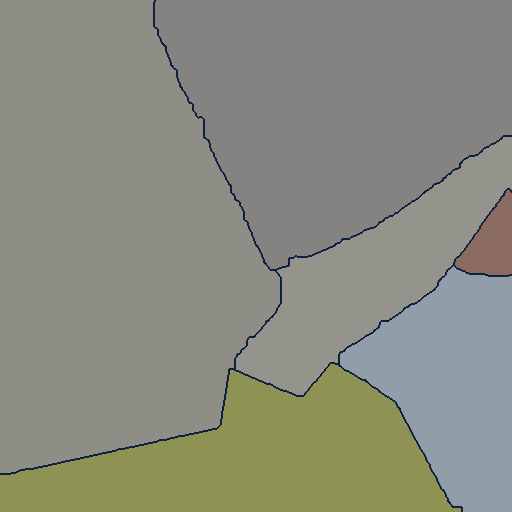}
    \end{subfigure}
    \hfill
    \begin{subfigure}[b]{0.16\textwidth}
        \includegraphics[width=\textwidth]{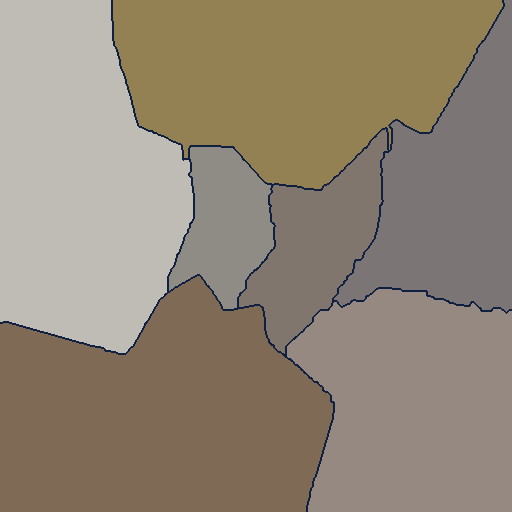}
    \end{subfigure}
    \hfill
    \begin{subfigure}[b]{0.16\textwidth}
        \includegraphics[width=\textwidth]{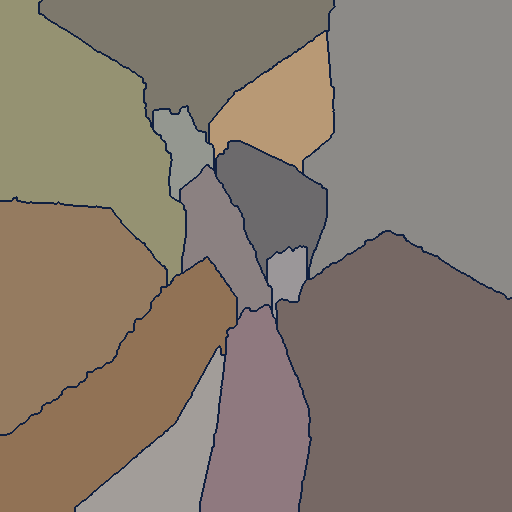}
    \end{subfigure}
    \hfill
    \begin{subfigure}[b]{0.16\textwidth}
        \includegraphics[width=\textwidth]{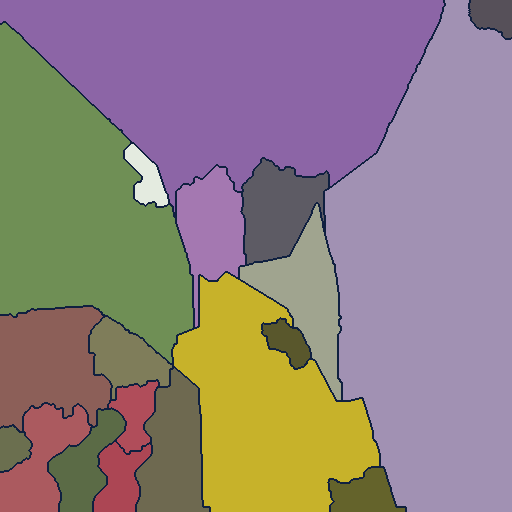}
    \end{subfigure}
    \hfill
    \begin{subfigure}[b]{0.16\textwidth}
        \includegraphics[width=\textwidth]{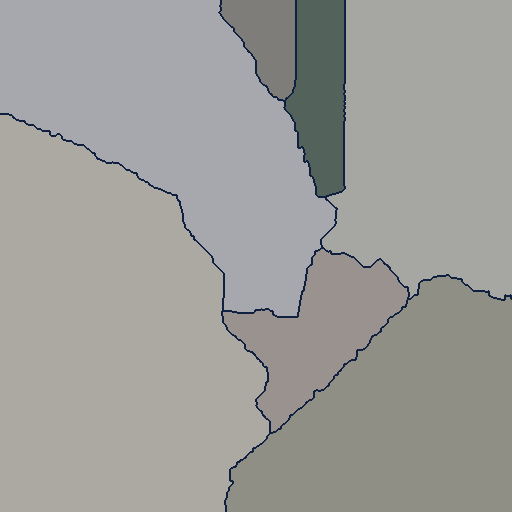}
    \end{subfigure}
    \caption{Exemplary segmentation results on the Prague texture segmentation dataset. \emph{From top to bottom:} input image, ground truth, FSEG \cite{yuan2015factorization}, PMCFA \cite{panagiotakis2011,panagiotakis2011slides}, PCA-MS \cite{mevenkamp2016} and the proposed method.}
    \label{fig:icpr2014_visual_results}
\end{figure*}

To assess the performance of our approach, we compare our results to several state-of-the-art algorithms that were used for segmentation on the Prague texture mosaics such as the Texel-based Segmentation (TS) \cite{todorovic2009}, Segmentation by Weighted Aggregation (SWA) \cite{galun2003}, Gaussian MRF Model With EM (GMRF) \cite{haindl2004}, 3-D Auto Regressive Model With EM (AR3D) \cite{haindl2006}, Texture Fragmentation and Reconstruction (TFR) and (TFR+) \cite{scarpa2009}, Regression-based Segmentation (RS) \cite{yuan2012image}, Factorization-Based Texture Segmentation 
(FSEG) \cite{yuan2015factorization}, Priority Multi-Class Flooding Algorithm (PMCFA) \cite{panagiotakis2011,panagiotakis2011slides} and Variational Multi-Phase Segmentation (PCA-MS) \cite{mevenkamp2016}.
\Idea{SUMMARIZE ALL METHODS THAT WE HAVEN'T MENTIONED IN RELATED WORK?}
Table~\ref{tbl:PragueColor_ICPR2014} provides the segmentation accuracy benchmark results as reported on the benchmark website \cite{haindl2008online} and in \cite{yuan2015factorization} as well as in \cite{mevenkamp2016}. In addition, Figure~\ref{fig:icpr2014_visual_results} depicts some of the segmentations produced by the four top-performing methods including our results for visual comparison. 
\Idea{DOES PMCFA USE IMPLICIT LIMITS ON THE NUMBER OF SEGMENTS? CONFIRM THIS!}

\subsection{Parameter sensitivity}

\begin{table*}
    \caption{Sensitivity w.r.t.\ the filter coherence penalty $\lambda$}
	\centering
	\begin{tabular}{@{} l *{8}{S[table-format=2.3,detect-inline-weight=math,table-align-text-post=false]} @{}}
		\toprule
		{$\lambda$}      & {0.1} & {1} & {5} & {10} & {15} & {20} & {50} & {100} \\
		\midrule
		$\uparrow$ CS    & 57.38 & 72.58 & 70.97 & 71.59 & 72.54 & 65.85 & 68.56 & 75.70 \\
		$\downarrow$ OS  & 9.67 & 18.64 & 18.69 & 18.73 & 18.63 & 18.71 & 18.67 & 14.81 \\
		$\downarrow$ US  & 34.65 & 7.98 & 3.66 & 7.78 & 3.67 & 14.58 & 11.72 & 7.98 \\
		$\downarrow$ ME  & 2.97 & 5.19 & 10.04 & 9.26 & 10.04 & 5.19 & 5.19 & 5.19 \\
		$\downarrow$ NE  & 0.42 & 5.97 & 10.42 & 10.13 & 10.41 & 6.14 & 6.23 & 5.96 \\
		$\downarrow$ O   & 26.14 & 9.44 & 12.12 & 12.97 & 10.04 & 11.57 & 11.36 & 9.34 \\
		$\downarrow$ C   & 26.22 & 31.05 & 31.07 & 31.19 & 31.03 & 31.84 & 33.88 & 31.12 \\
		$\uparrow$ CA    & 68.49 & 79.52 & 78.90 & 79.59 & 79.75 & 75.48 & 77.23 & 81.27 \\
		$\uparrow$ CO    & 78.31 & 83.99 & 83.13 & 84.34 & 84.14 & 81.47 & 82.55 & 85.80 \\
		$\uparrow$ CC    & 74.22 & 90.96 & 91.19 & 90.45 & 90.77 & 86.06 & 88.20 & 90.96 \\
		$\downarrow$ I.  & 21.69 & 16.01 & 16.87 & 15.66 & 15.86 & 18.53 & 17.45 & 14.20 \\
		$\downarrow$ II. & 5.00 & 1.49 & 1.38 & 1.54 & 1.52 & 1.99 & 1.88 & 1.52 \\
		$\uparrow$ EA    & 74.10 & 85.14 & 84.67 & 85.41 & 85.37 & 81.54 & 83.22 & 86.58 \\
		$\uparrow$ MS    & 69.46 & 79.69 & 79.14 & 79.79 & 79.78 & 75.78 & 77.27 & 81.47 \\
		$\downarrow$ RM  & 7.32 & 3.85 & 3.83 & 3.79 & 3.73 & 4.71 & 4.09 & 3.59 \\
		$\uparrow$ CI    & 75.11 & 86.22 & 85.81 & 86.33 & 86.33 & 82.58 & 84.23 & 87.42 \\
		$\downarrow$ GCE & 6.34 & 8.87 & 8.69 & 9.86 & 8.80 & 9.58 & 9.78 & 8.77 \\
		$\downarrow$ LCE & 5.31 & 5.90 & 5.99 & 6.83 & 6.12 & 5.95 & 6.05 & 5.78 \\
		$\downarrow$ dD  & 12.76 & 9.99 & 10.40 & 10.21 & 9.93 & 11.19 & 10.71 & 9.06 \\
		$\downarrow$ dM  & 11.19 & 5.39 & 5.48 & 5.47 & 5.31 & 6.10 & 5.87 & 4.96 \\
		$\downarrow$ dVI & 14.02 & 15.37 & 15.50 & 15.27 & 15.33 & 15.15 & 15.28 & 15.17 \\
		\bottomrule
	\end{tabular}
	
	\label{tbl:paramsense_filter_coherence}
\end{table*}

\begin{table*}
	\centering
	\caption{Sensitivity w.r.t.\ the number of filters $K$}
	\begin{tabular}{@{} l *{8}{S[table-format=2.3,detect-inline-weight=math,table-align-text-post=false]} @{}}
		\toprule
		{K}              & {11}  & {21}  & {31}  & {41}  & {61}  & {81}  & {122} & {162} \\
		\midrule
		$\uparrow$ CS    & 32.37 & 52.23 & 70.25 & 72.59 & 59.13 & 43.36 & 17.43 & 0.47 \\
		$\downarrow$ OS  & 6.93 & 13.94 & 15.13 & 18.73 & 14.74 & 16.10 & 21.06 & 0.00 \\
		$\downarrow$ US  & 66.81 & 28.78 & 13.65 & 7.78 & 8.49 & 3.65 & 30.43 & 46.97 \\
		$\downarrow$ ME  & 0.15 & 14.43 & 11.91 & 5.19 & 16.14 & 35.38 & 29.49 & 51.15 \\
		$\downarrow$ NE  & 0.00 & 14.29 & 12.53 & 6.18 & 15.99 & 35.68 & 27.38 & 47.06 \\
		$\downarrow$ O   & 51.79 & 42.58 & 14.76 & 10.17 & 17.22 & 27.10 & 52.82 & 76.49 \\
		$\downarrow$ C   & 36.68 & 31.94 & 22.52 & 31.26 & 31.78 & 46.46 & 76.02 & 64.75 \\
		$\uparrow$ CA    & 44.39 & 61.58 & 75.33 & 80.02 & 71.65 & 63.04 & 38.22 & 21.89 \\
		$\uparrow$ CO    & 59.30 & 71.38 & 82.67 & 84.27 & 77.58 & 70.48 & 49.76 & 38.53 \\
		$\uparrow$ CC    & 46.42 & 68.23 & 81.84 & 91.25 & 88.68 & 84.58 & 68.67 & 42.50 \\
		$\downarrow$ I.  & 40.70 & 28.62 & 17.33 & 15.73 & 22.42 & 29.52 & 50.24 & 61.47 \\
		$\downarrow$ II. & 14.82 & 6.25 & 2.83 & 1.38 & 2.16 & 3.69 & 9.53 & 15.08 \\
		$\uparrow$ EA    & 49.77 & 67.15 & 80.61 & 85.61 & 78.86 & 71.84 & 49.06 & 31.98 \\
		$\uparrow$ MS    & 39.39 & 58.60 & 75.53 & 80.28 & 72.36 & 62.84 & 32.61 & 11.44 \\
		$\downarrow$ RM  & 17.04 & 10.77 & 5.81 & 3.82 & 5.27 & 6.13 & 14.13 & 18.71 \\
		$\uparrow$ CI    & 51.20 & 68.39 & 81.38 & 86.62 & 80.79 & 74.35 & 53.51 & 35.65 \\
		$\downarrow$ GCE & 3.21 & 7.71 & 7.75 & 8.65 & 12.27 & 17.57 & 19.34 & 25.64 \\
		$\downarrow$ LCE & 2.27 & 5.35 & 6.46 & 5.88 & 7.88 & 12.06 & 14.01 & 17.93 \\
		$\downarrow$ dD  & 21.09 & 16.48 & 11.23 & 9.80 & 14.39 & 19.70 & 30.97 & 37.73 \\
		$\downarrow$ dM  & 38.36 & 18.27 & 7.81 & 5.28 & 7.77 & 11.91 & 31.97 & 43.74 \\
		$\downarrow$ dVI & 11.81 & 13.52 & 14.41 & 15.34 & 15.85 & 16.05 & 15.42 & 13.46 \\
		\bottomrule
	\end{tabular}
	\label{tbl:paramsense_filter_count}
\end{table*}

We explore the sensitivity of our method with respect to the most influential parameters. 
To that end, we conduct an evaluation of our method with varying parameters on a representative subset of images from the Prague benchmark dataset drawn from the different categories all, bark, flowers, glass, nature, stone and textile.
We examine the filter size, the number of learned filters $K$, the weight of the filter coherence penalty $\lambda,$ and the parameters $\mu$ and $\nu$ of the employed non-linearities. We vary each of them while keeping the others fixed at the values described in Subsection~\ref{subsec:experiments_prague}.

We start with the parameter $\lambda$ which controls the maximum coherence between all filter pairs and which is given in Eq.~\eqref{eq:final_learning_problem}.
Table~\ref{tbl:paramsense_filter_coherence} shows that if $\lambda$ is close to zero which effectively disables this constraint, segmentation results deteriorate significantly. For $\lambda$ larger than $1$, we observe only negligible changes in segmentation results across all quality metrics. These results underline the importance of the constraint in our learning objective but also reveal that the choice of its exact value is not critical as long as it is large enough.

Next, we investigate the influence of the number of filters $K$.
We conclude from Table~\ref{tbl:paramsense_filter_count} that the segmentation quality increases for up to $41$ filters and deteriorates for larger numbers. The initial improvement might be explained by the increased discriminatory power obtained from a larger number of different filters.
The deterioration of the quality for a larger number of filters might be explained by an over-segmentation caused by irrelevant features.

We continue by studying the influence of the filter size.
The choice of the filter size should relate to the scale of the texture.
Although the Prague texture mosaics expose a relatively large variety of texture scales, we find that filter sizes of $7$ and $9$ pixels achieve the best results in average (see Table \ref{tbl:paramsense_filter_size}), which confirms the choice of other works, e.g.\ \cite{mevenkamp2016, yuan2015factorization}. For small filter sizes, within-texture variations are similar to variations at texture boundaries which leads to undersegmentation in the segmentation stage. 
For our method, we also observe that large filters lead to a decreased localization of texture boundaries, and to larger spurious segments at texture boundaries as depicted in Figure~\ref{fig:region_merging}.  
Now we consider the non-linearity parameter $\nu$
which we used for relaxation of the $\ell_0$ jump penalty for filter learning in Eq.~\eqref{eq:l0_log_approx}.
We recall that for large $\nu$ 
the surrogate function  approximates the original sparsifying function well \cite{Kiechle2013,nieuwenhuis2013}.
Table~\ref{tbl:paramsense_sparsity_nu} lists segmentation results over a large range of $\nu$.
The segmentation fails for small values of $\nu$ and improves when increasing it. Due to the decreasing slope of the surrogate function for large values of $\nu$, the learning algorithm converges more slowly. Our choice in the experiments reflects a trade-off between convergence speed and approximation accuracy.

Finally, we investigate the sensitivity of the corresponding parameter $\mu$ in the non-linearity $\sigma$ of Eq.~\eqref{eq:sparsity_objective} in Table~\ref{tbl:paramsense_nonlin}. We find that the overall segmentation results are robust over a large range of choices of $\mu$ and segmentation quality only starts suffering for very large values of $\mu$ where the shape of $\sigma$ degenerates.

\begin{table}
	\centering
	\caption{Sensitivity w.r.t.\ the filter size}
	\begin{tabular}{@{} l *{5}{S[table-format=2.3,detect-inline-weight=math,table-align-text-post=false]} @{}}
		\toprule
		{filter size} & {5} & {7} & {9} & {11} & {13}   \\
		\midrule
		$\uparrow$ CS    & 55.23 & 69.74 & 71.59 & 59.68 & 54.99 \\
		$\downarrow$ OS  & 9.22 & 14.05 & 18.73 & 13.06 & 8.56 \\
		$\downarrow$ US  & 39.69 & 14.28 & 7.78 & 16.11 & 9.31 \\
		$\downarrow$ ME  & 6.63 & 8.39 & 9.26 & 17.14 & 25.10 \\
		$\downarrow$ NE  & 5.93 & 7.22 & 10.13 & 17.95 & 25.22 \\
		$\downarrow$ O   & 38.39 & 6.18 & 12.97 & 12.81 & 21.91 \\
		$\downarrow$ C   & 21.64 & 16.97 & 31.19 & 33.25 & 32.46 \\
		$\uparrow$ CA    & 62.27 & 78.59 & 79.59 & 71.45 & 66.58 \\
		$\uparrow$ CO    & 73.21 & 84.33 & 84.34 & 78.28 & 74.54 \\
		$\uparrow$ CC    & 64.40 & 87.46 & 90.45 & 86.10 & 81.01 \\
		$\downarrow$ I.  & 26.79 & 15.67 & 15.66 & 21.72 & 25.46 \\
		$\downarrow$ II. & 8.39 & 1.66 & 1.54 & 2.30 & 3.55 \\
		$\uparrow$ EA    & 66.87 & 83.50 & 85.41 & 78.82 & 74.44 \\
		$\uparrow$ MS    & 60.30 & 78.59 & 79.79 & 72.27 & 66.01 \\
		$\downarrow$ RM  & 10.50 & 5.42 & 3.79 & 5.64 & 6.31 \\
		$\uparrow$ CI    & 67.78 & 84.57 & 86.33 & 80.37 & 76.00 \\
		$\downarrow$ GCE & 4.07 & 7.60 & 9.86 & 12.27 & 15.64 \\
		$\downarrow$ LCE & 3.59 & 4.80 & 6.83 & 8.01 & 11.30 \\
		$\downarrow$ dD  & 14.56 & 9.97 & 10.21 & 14.00 & 17.09 \\
		$\downarrow$ dM  & 21.38 & 5.96 & 5.47 & 7.75 & 10.35 \\
		$\downarrow$ dVI & 12.90 & 14.79 & 15.27 & 15.58 & 15.51 \\
		\bottomrule
	\end{tabular}
	\label{tbl:paramsense_filter_size}
\end{table}

\begin{table}
	\centering
	\caption{Sensitivity w.r.t.\ the non-linearity parameter $\nu$}
	\begin{tabular}{@{} l *{5}{S[table-format=2.3,detect-inline-weight=math,table-align-text-post=false]} @{}}
		\toprule
		{$\nu$} 		 & {100} & {500} & {1000} & {2000} & {3000}   \\
		\midrule
		$\uparrow$ CS    & 0.00 & 29.57 & 49.31 & 71.56 & 69.02 \\
		$\downarrow$ OS  & 0.00 & 0.00 & 6.83 & 18.71 & 23.61 \\
		$\downarrow$ US  & 99.93 & 69.31 & 47.39 & 7.69 & 7.99 \\
		$\downarrow$ ME  & 0.00 & 0.00 & 0.00 & 10.04 & 3.10 \\
		$\downarrow$ NE  & 0.00 & 0.00 & 0.00 & 10.42 & 3.86 \\
		$\downarrow$ O   & 100.00 & 58.36 & 44.93 & 10.50 & 13.68 \\
		$\downarrow$ C   & 71.63 & 40.48 & 46.06 & 31.09 & 32.54 \\
		$\uparrow$ CA    & 8.43 & 38.42 & 53.34 & 78.44 & 77.09 \\
		$\uparrow$ CO    & 28.05 & 54.02 & 64.42 & 83.80 & 81.35 \\
		$\uparrow$ CC    & 8.44 & 39.29 & 57.00 & 87.97 & 91.25 \\
		$\downarrow$ I.  & 71.95 & 45.98 & 35.58 & 16.20 & 18.65 \\
		$\downarrow$ II. & 24.62 & 16.39 & 13.10 & 2.65 & 1.40 \\
		$\uparrow$ EA    & 12.80 & 43.06 & 57.62 & 84.16 & 83.57 \\
		$\uparrow$ MS    & -7.92 & 31.03 & 47.78 & 78.78 & 77.82 \\
		$\downarrow$ RM  & 30.45 & 19.14 & 15.15 & 4.40 & 4.36 \\
		$\uparrow$ CI    & 15.25 & 44.71 & 59.04 & 84.97 & 84.86 \\
		$\downarrow$ GCE & 0.14 & 2.16 & 3.13 & 8.58 & 9.00 \\
		$\downarrow$ LCE & 0.14 & 1.78 & 2.55 & 6.04 & 6.20 \\
		$\downarrow$ dD  & 36.01 & 23.55 & 18.61 & 10.04 & 11.23 \\
		$\downarrow$ dM  & 75.36 & 43.91 & 35.86 & 7.01 & 5.93 \\
		$\downarrow$ dVI & 9.14 & 11.20 & 12.43 & 15.02 & 15.74 \\
		\bottomrule
	\end{tabular}
	\label{tbl:paramsense_sparsity_nu}
\end{table}

\begin{table}
	\caption{Sensitivity w.r.t.\ non-linearity parameter $\mu$}
	\centering
	\begin{tabular}{@{} l *{5}{S[table-format=2.3,detect-inline-weight=math,table-align-text-post=false]} @{}}
		\toprule
		{$\mu$}          & {10} & {100} & {2000} & {5000} & {10000} \\
		\midrule
		$\uparrow$ CS    & 74.69 & 71.79 & 72.59 & 65.99 & 24.30 \\
		$\downarrow$ OS  & 18.44 & 22.18 & 18.73 & 15.33 & 6.80 \\
		$\downarrow$ US  & 7.97 & 7.89 & 7.78 & 6.89 & 36.01 \\
		$\downarrow$ ME  & 0.00 & 0.00 & 5.19 & 16.82 & 29.83 \\
		$\downarrow$ NE  & 0.66 & 0.75 & 6.18 & 16.76 & 29.64 \\
		$\downarrow$ O   & 13.33 & 13.96 & 10.17 & 15.18 & 59.31 \\
		$\downarrow$ C   & 31.08 & 31.16 & 31.26 & 30.92 & 51.44 \\
		$\uparrow$ CA    & 79.50 & 79.04 & 80.02 & 72.85 & 41.24 \\
		$\uparrow$ CO    & 83.65 & 82.92 & 84.27 & 79.35 & 53.97 \\
		$\uparrow$ CC    & 91.40 & 91.46 & 91.25 & 84.99 & 60.72 \\
		$\downarrow$ I.  & 16.35 & 17.08 & 15.73 & 20.65 & 46.03 \\
		$\downarrow$ II. & 1.43 & 1.39 & 1.38 & 2.73 & 10.29 \\
		$\uparrow$ EA    & 84.96 & 84.83 & 85.61 & 78.93 & 49.49 \\
		$\uparrow$ MS    & 80.21 & 79.55 & 80.28 & 72.51 & 36.61 \\
		$\downarrow$ RM  & 2.95 & 3.79 & 3.82 & 5.22 & 14.07 \\
		$\uparrow$ CI    & 86.12 & 85.93 & 86.62 & 80.42 & 52.84 \\
		$\downarrow$ GCE & 8.22 & 8.12 & 8.65 & 13.34 & 15.58 \\
		$\downarrow$ LCE & 5.60 & 5.83 & 5.88 & 8.56 & 12.02 \\
		$\downarrow$ dD  & 9.85 & 10.19 & 9.80 & 13.50 & 27.86 \\
		$\downarrow$ dM  & 5.31 & 5.41 & 5.28 & 8.21 & 29.43 \\
		$\downarrow$ dVI & 15.59 & 15.57 & 15.34 & 15.26 & 14.53 \\
		\bottomrule
	\end{tabular}
	\label{tbl:paramsense_nonlin}
\end{table}

\subsection{Histology dataset}
\begin{figure*}[!htbp]
    \centering
    \begin{subfigure}[b]{0.16\textwidth}
        \includegraphics[width=\textwidth]{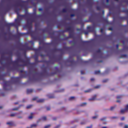}
    \end{subfigure}
    \hfill
    \begin{subfigure}[b]{0.16\textwidth}
        \includegraphics[width=\textwidth]{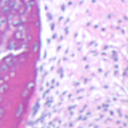}
    \end{subfigure}
    \hfill
    \begin{subfigure}[b]{0.16\textwidth}
        \includegraphics[width=\textwidth]{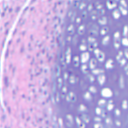}
    \end{subfigure}
    \hfill
    \begin{subfigure}[b]{0.16\textwidth}
        \includegraphics[width=\textwidth]{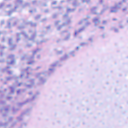}
    \end{subfigure}
    \hfill
    \begin{subfigure}[b]{0.16\textwidth}
        \includegraphics[width=\textwidth]{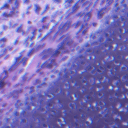}
    \end{subfigure}
    \hfill
    \begin{subfigure}[b]{0.16\textwidth}
        \includegraphics[width=\textwidth]{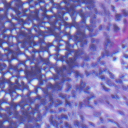}
    \end{subfigure}
    \\
    \vspace{1mm}
    \begin{subfigure}[b]{0.16\textwidth}
        \includegraphics[width=\textwidth]{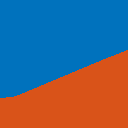}
    \end{subfigure}
    \hfill
    \begin{subfigure}[b]{0.16\textwidth}
        \includegraphics[width=\textwidth]{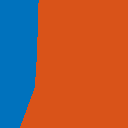}
    \end{subfigure}
    \hfill
    \begin{subfigure}[b]{0.16\textwidth}
        \includegraphics[width=\textwidth]{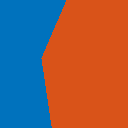}
    \end{subfigure}
    \hfill
    \begin{subfigure}[b]{0.16\textwidth}
        \includegraphics[width=\textwidth]{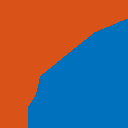}
    \end{subfigure}
    \hfill
    \begin{subfigure}[b]{0.16\textwidth}
        \includegraphics[width=\textwidth]{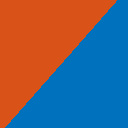}
    \end{subfigure}
    \hfill
    \begin{subfigure}[b]{0.16\textwidth}
        \includegraphics[width=\textwidth]{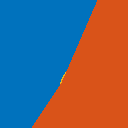}
    \end{subfigure}
    \\
    \vspace{1mm}
    \begin{subfigure}[b]{0.16\textwidth}
        \includegraphics[width=\textwidth]{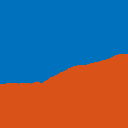}
    \end{subfigure}
    \hfill
    \begin{subfigure}[b]{0.16\textwidth}
        \includegraphics[width=\textwidth]{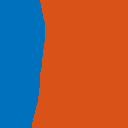}
    \end{subfigure}
    \hfill
    \begin{subfigure}[b]{0.16\textwidth}
        \includegraphics[width=\textwidth]{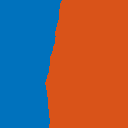}
    \end{subfigure}
    \hfill
    \begin{subfigure}[b]{0.16\textwidth}
        \includegraphics[width=\textwidth]{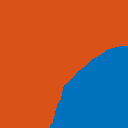}
    \end{subfigure}
    \hfill
    \begin{subfigure}[b]{0.16\textwidth}
        \includegraphics[width=\textwidth]{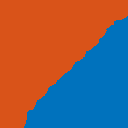}
    \end{subfigure}
    \hfill
    \begin{subfigure}[b]{0.16\textwidth}
        \includegraphics[width=\textwidth]{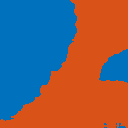}
    \end{subfigure}
    \\
    \vspace{1mm}
    \begin{subfigure}[b]{0.16\textwidth}
        \includegraphics[width=\textwidth]{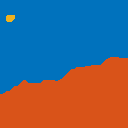}
    \end{subfigure}
    \hfill
    \begin{subfigure}[b]{0.16\textwidth}
        \includegraphics[width=\textwidth]{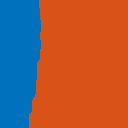}
    \end{subfigure}
    \hfill
    \begin{subfigure}[b]{0.16\textwidth}
        \includegraphics[width=\textwidth]{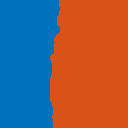}
    \end{subfigure}
    \hfill
    \begin{subfigure}[b]{0.16\textwidth}
        \includegraphics[width=\textwidth]{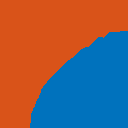}
    \end{subfigure}
    \hfill
    \begin{subfigure}[b]{0.16\textwidth}
        \includegraphics[width=\textwidth]{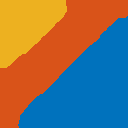}
    \end{subfigure}
    \hfill
    \begin{subfigure}[b]{0.16\textwidth}
        \includegraphics[width=\textwidth]{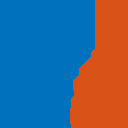}
    \end{subfigure}
    \caption{Segmentation results on the histology dataset from \cite{mccann2014}. \emph{From top to bottom:} input image, ground truth,  ORTSEG \cite{mccann2014}, and the proposed method.}
    \label{fig:histo_visual_results}
\end{figure*}

We apply our method to the histology dataset used in \cite{mccann2014}. The dataset contains 36 color images of size $128\times 128$ pixels of stained tissue along with segmentations by an expert. Instead of the adaptive color quantization used in \cite{mccann2014}, we simply convert the image to gray-scale prior to processing. Since the images are considerably smaller than the Prague texture mosaics, we reduce the filter size to $5\times 5$, learn only 13 filters and therefore adjust the trade-off parameter in the  segmentation stage to a fixed $\gamma=0.8$ but otherwise use the same setup as in the Prague texture experiment. 
We point out that switching to this quite different class of images
only required the adjustment of these few parameters.
The learning stage requires approximately $2$~minutes
and the segmentation stage around $3$~seconds.
Some of the results
are given in Figure~\ref{fig:histo_visual_results}.

\subsection{Discussion}

From Table~\ref{tbl:PragueColor_ICPR2014}
we observe
that the proposed method significantly improves upon
most existing approaches in the Prague texture segmentation benchmark. 
Moreover, the segmentations obtained by the proposed method are competitive 
with the previously best performing method PMCFA. 
PMCFA and the proposed method yield a comparable number of first and second ranks.
   The segmentation examples in Figure~\ref{fig:icpr2014_visual_results}
 indicate that our method gives very satisfactory results for 
 segments with clear repeated texture patterns, as for instance in the first three examples. 
 Erroneous segmentations appear mostly 
 when quite different patterns 
 such as the red blossoms on green background in the fifth image
 are present in a segment.
 A possible cause for this is that the blossoms are interpreted as a texture on its own on a smaller scale.
 Qualitatively, we observe that our method tends to a slight oversegmentation 
  when large color contrasts are present. This is not the case for PMCFA.
  Compared to PMCFA, the proposed method produces smoother boundaries.
 
In addition to the Prague texture segmentation benchmark,
the algorithm produces useful segmentations of the tissue images of \cite{mccann2014}.
It is mostly very close to the expert annotations. 
We stress that we only had to adapt the maximum number of filters $K,$ their maximum size and the segmentation hyperparameter $\gamma$ 
to obtain the presented results.
This indicates the potential of the proposed method for segmenting different classes of images.

The main trade-off of our method is a relatively long processing time per image.
In contrast to most other methods where a fixed set of features is used for segmentation, we here run the learning stage prior to segmentation, which increases the overall running time.
The computational complexity of the learning stage is primarily determined by size and number of the filters as well as the number of patches and their channels that are used for learning.
A speed-up could be achieved by 
reducing the number of training samples. We observed that reducing the number of samples for training from $50K$ to $10K$ only slightly decreased the segmentation quality.
Also, filter learning is so far started with a random initialization. In practical applications, the filter set can be initialized with prelearned filters which could bring down the required number of iterations during learning and therefore significantly decrease the overall running time. A further speed-up might be obtained by an optimized implementation.

\section{Conclusion}

We have developed a method
for unsupervised texture segmentation 
where the features are learned from images without ground truth segmentation.
Our first main contribution was the development of a corresponding model
based on local homogeneity assumptions.
We learn convolutional features in a way that they produce approximately piecewise constant feature images and combine this with the piecewise constant Mumford-Shah model.
Our second main contribution was the development of 
a practical algorithm for unsupervised texture segmentation based on that model.
To make the problem computationally tractable,
we relaxed it
and we decomposed it into a filter learning stage and a segmentation stage.
In the filter learning stage, we employed a geometric conjugate gradient descent method,
whereas in the segmentation stage,
we used the Lagrange formulation of the piecewise constant Mumford-Shah model 
which we proposed to augment with a Mahalanobis distance as data term.
The proposed algorithm 
yields competitive results on the standard benchmark dataset
for unsupervised texture segmentation. 
Furthermore, switching to the quite different class of histological images
only required the adjustment of a few parameters.
The improved segmentation quality underpins the idea of learning features adapted to the image under consideration. The proposed approach may be especially valuable in situations where creating large training sets of accurate ground truth segmentations or hand-crafting features is expensive.

Topics of future research include speeding-up the proposed method as explained in the discussion section as well as approaching the proposed non-smooth model more directly, that is, employing fewer relaxation steps.

\appendix

We derive the Euclidean gradient required in the numerical optimization of the filter learning problem described in Eq.~\eqref{eq:learning_function}.

The cost function to minimize in the learning stage consists of three terms, one each for the approximated cost of the jump set $f(\AOF{})$, the centroid penalty $r(\AOF{})$ and the coherence penalty $h(\AOF{})$.

\vspace{2mm}
\noindent
\textbf{Sparsity objective:}
First, we provide the derivative of the approximated cost of the jump set. Explicitly, let $\AOF{k} \in \R{n}$ a vectorized 2D filter of size $\sqrt{n} \times \sqrt{n}$ that is applied on a set of $M$ vectorized 2D image patches $\Matrix{U} \in \R{n \times M}$ by taking their standard inner product $\AOF{k}^{\top} \Vector{U} \in \R{1\times M}$ and for a set of filters we get $\AOF{} \Matrix{U} \in \R{K \times M}$ accordingly.
By denoting $\mathbf{D}_{a_s} = \nabla_{a_s} \sigma (\boldsymbol{\Phi} \mathbf{U})$ shorthand for the difference of features along $\Vector{a}_s$ and using $\odot$ for the Hadamard product, we obtain
\begin{multline} \label{eq:gradient_sparsity_objective}
\frac{\partial}{\partial \AOF{}} f(\AOF{}) \\
= 4 \nu \mu \sum_{s=1} \omega_{a_s} \bigg [ \left(\frac{\mathbf{D}_{a_s}}{1+\nu \Vert \mathbf{D}_{a_s} \Vert^2} \odot \frac{ \boldsymbol{\Phi} \mathbf{U}_{a_s} }{ 1 + \mu (\boldsymbol{\Phi} \mathbf{U}_{a_s})^2 } \right) \mathbf{U}_{a_s}^{\top}\\
- \left( \frac{\mathbf{D}_{a_s}}{1+\nu \Vert \mathbf{D}_{a_s} \Vert^2} \odot \frac{ \boldsymbol{\Phi} \mathbf{U}_{0} }{ 1 + \mu (\boldsymbol{\Phi} \mathbf{U}_{0})^2 } \right) \mathbf{U}_{0}^{\top} \bigg ]
\end{multline}
for the derivative of $f$.
Here, $\mathbf{U}_{a_s}$ is the $n \times M$ data matrix containing vectorized patches $\mathbf{U}_i^{a_s}$ cropped from the $M$ sampled super-patches in direction $\Vector{a}_s$ according to Figure~\ref{fig:patch_crops}.

\vspace{2mm}
\noindent
\textbf{Centroid penalty:}
Second, we require the derivative of the centroid constraint Eq.~\eqref{eq:centered_moment_penalty}.
This constraint acts on each filter independently. For the individual filter, we get
\begin{multline*}
\frac{\partial}{\partial \AOF{k}} h(\AOF{}) = \\
\frac{4}{w_{-}} \left[ \frac{c_{k,x} \Matrix{P}_x}{1-c_{k,x}^2} + \frac{c_{k,y} \Matrix{P}_y}{1-c_{k,y}^2} + \frac{1}{2} (c_{k,x}-c_{k,y}) (\Matrix{P}_x-\Matrix{P}_y) \right] \AOF{k}\end{multline*}
by using $w_{-} = \frac{\sqrt{n}-1}{2}$ as a shorthand notation for the half width of the filter.
By stacking the individual derivatives we can then write the derivative of $h$ with respect to the entire filter set as
\begin{equation} \label{eq:centroid_derivative}
\frac{\partial}{\partial \AOF{}} h(\AOF{}) = \left[ \frac{\partial}{\partial \AOF{1}} h(\AOF{}), \dots, \frac{\partial}{\partial \AOF{K}} h(\AOF{})  \right]^{\top}.
\end{equation}

\vspace{2mm}
\noindent
\textbf{Coherence penalty:}
Last, the gradient of the coherence penalty Eq.~\eqref{eq:incoherence_penalty} is given in \cite{hawe2013} and reads as
\begin{equation} \label{eq:derivative_coherence}
\frac{\partial}{\partial \AOF{}} r(\AOF{}) = \left[ \sum_{1\leq i < j \leq k} \frac{2 \AOF{i}^{\top} \AOF{j}}{1 - \left( \AOF{i}^{\top} \AOF{j} \right)^2} \left( \mathcal{E}_{ij} + \mathcal{E}_{ji} \right) \right] \AOF{}.
\end{equation}
Here $\mathcal{E}_{ij}$ is a matrix with a one in component $ij$ and zero elsewhere.

\vspace{2mm}
\noindent
Finally, we obtain the gradient of the cost function Eq.~\eqref{eq:learning_function} by combining Eq.~\eqref{eq:gradient_sparsity_objective}, Eq.~\eqref{eq:centroid_derivative} and Eq.~\eqref{eq:derivative_coherence}
\begin{equation}
\nabla E(\AOF{}) = \frac{\partial}{\partial \AOF{}} f(\AOF{}) + \lambda \frac{\partial}{\partial \AOF{}} r(\AOF{}) + \kappa \frac{\partial}{\partial \AOF{}} h(\AOF{}). \nonumber
\end{equation}

\section*{Acknowledgment}
This work has been supported by the German Research Foundation (DFG) through grants number KL 2189/9-1,
STO1126/2-1, and  WE5886/4-1.

\ifCLASSOPTIONcaptionsoff
  \newpage
\fi

{\footnotesize
\bibliographystyle{IEEEtran}
\bibliography{references}
}

\begin{IEEEbiography}[{\includegraphics[width=1in,height=1.25in,clip,keepaspectratio]{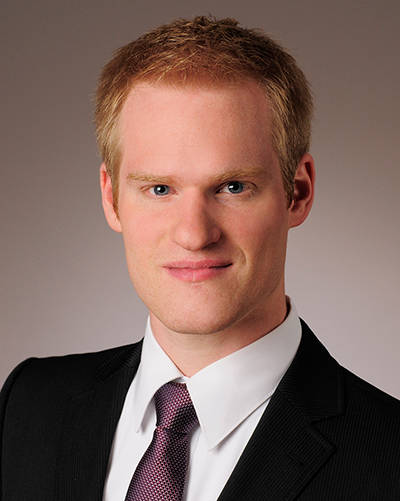}}]{Martin Kiechle}
received his Diploma degree in Electrical and Computer Engineering and his Honours degree in Technology Management both from Technische Universit\"at M\"unchen in 2012. Since 2013, he has been working as a researcher in the Research Group for Geometric Optimization and Machine Learning at Technische Universit\"at M\"unchen. His research interests include signal and image processing, representation learning, geometric optimization, and inverse problems.
\end{IEEEbiography}

\begin{IEEEbiography}[{\includegraphics[width=1in,height=1.25in,clip,keepaspectratio]{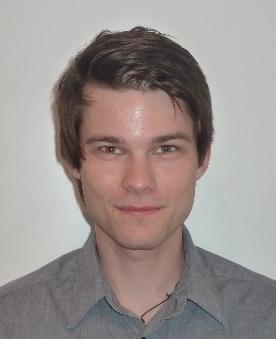}}]{Martin Storath}
received his Diploma degree in Mathematics  in 2008, his Honours degree in Technology Management in 2009, and his doctoral degree 
in Mathematics in 2013,
all from Technische Universit\"at M\"unchen.
From 2010 to 2013, he worked as researcher at the Institute of Biomathematics, Helmholtz Zentrum M\"unchen.
From 2013 to 2016, he was a post-doctoral researcher at the Biomedical Imaging Group, \'Ecole Polytechnique
F\'ed\'erale de Lausanne, Switzerland. 
Currently, he is a postdoc at the Image Analysis and Learning Group, Universit\"at Heidelberg.
His research interests include signal and image processing, biomedical imaging,
variational methods, and inverse problems.
\end{IEEEbiography}

\begin{IEEEbiography}[{\includegraphics[width=1in,height=1.25in,clip,keepaspectratio]{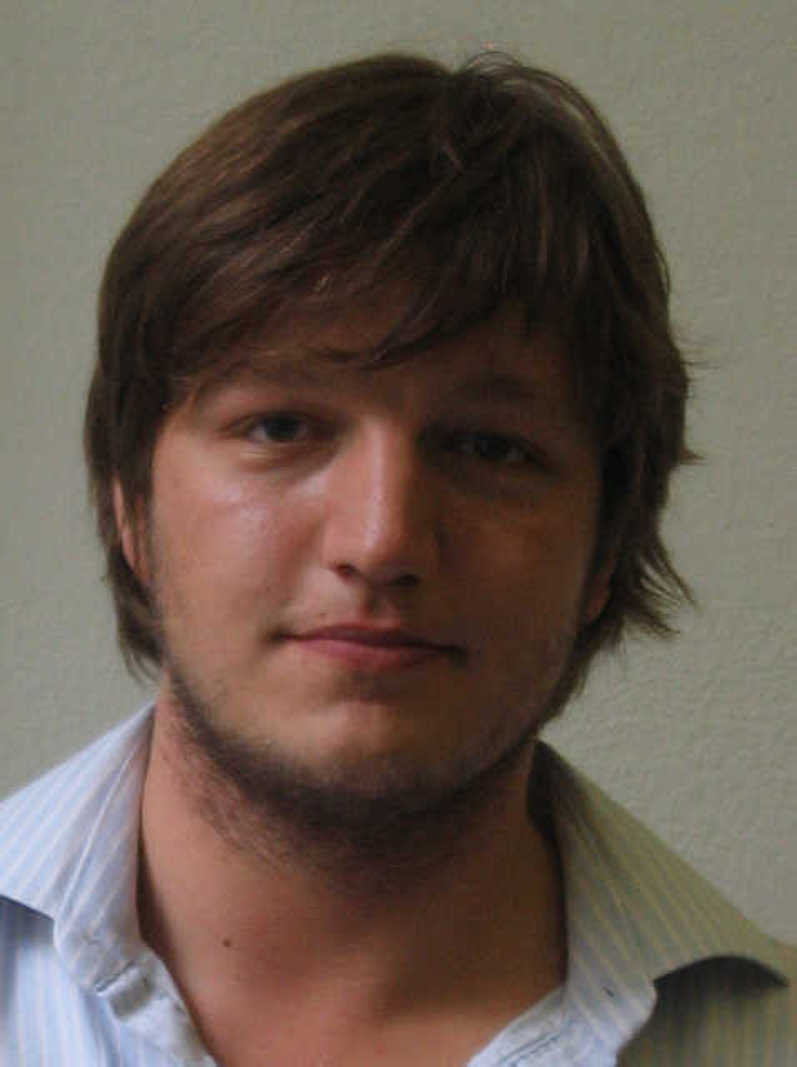}}]{Andreas Weinmann}
	received his Diploma degree in mathematics and computer science from Technische Universit\"at M\"unchen in 2006
	and his Ph.D. degree from Technische Universit\"at Graz in 2010 (both with highest distinction).
	Currently he is affiliated with the Helmholtz Center Munich and the  
	mathematics department at TU Munich and the Darmstadt University of Applied Sciences.
	His research interests are applied analysis as well as signal and image processing.
\end{IEEEbiography}

\begin{IEEEbiography}[{\includegraphics[width=1in,height=1.25in,clip,keepaspectratio]{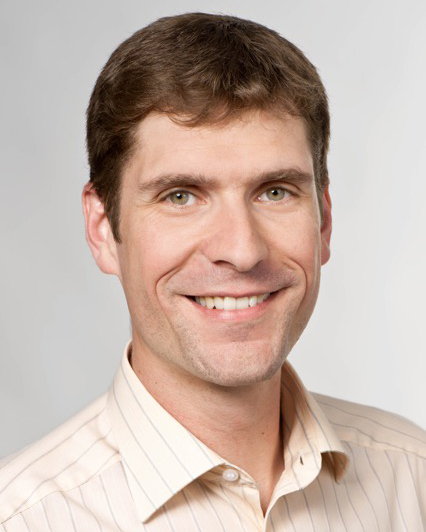}}]{Martin Kleinsteuber}
received his Ph.D. in Mathematics from the University of Würzburg, Germany, in 2006. After post-doc positions at National ICT Australia Ltd., the Australian National University, Canberra, Australia, and the University of Würzburg, he has been appointed assistant professor for geometric optimization and machine learning at the Department of Electrical and Computer Engineering, TU München, Germany, in 2009. He won the SIAM student paper prize in 2004 and the Robert-Sauer-Award of the Bavarian Academy of Science in 2008 for his works on Jacobi-type methods on Lie algebras. Since 2016, he is leading the Data Science Group at Mercateo AG, Munich.
\end{IEEEbiography}

\end{document}